\pdfoutput=1
\documentclass[]{article}
\usepackage[square,sort,comma,numbers]{natbib}
\usepackage[final]{nips_2016}

\usepackage[utf8]{inputenc} % allow utf-8 input
\usepackage[T1]{fontenc}    % use 8-bit T1 fonts
\usepackage[bookmarks, colorlinks=true, citecolor=Violet,linkcolor=Mahogany,urlcolor=blue]{hyperref}
\usepackage{booktabs}       % professional-quality tables
\usepackage{amsfonts}       % blackboard math symbols
\usepackage{nicefrac}       % compact symbols for 1/2, etc.
\usepackage{microtype}      % microtypography
\usepackage{times}
\usepackage{epsfig}
\usepackage{wrapfig}
\usepackage{float}
\usepackage{amsmath}
\usepackage{amssymb}
\usepackage{amsthm}
\usepackage{rotating}
\usepackage{algorithm}
\usepackage{algorithmic}
\usepackage{multirow}
\usepackage{multicol}
\usepackage{blindtext}
\usepackage{titlesec}
\usepackage[usenames,dvipsnames,dvinames]{xcolor}
\newtheorem{lemma}{Lemma}
\newtheorem{theorem}{Theorem}
\newtheorem{definition}{Definition}
\newtheorem{proposition}{Proposition}

\DeclareMathOperator*{\argmax}{argmax}
\DeclareMathOperator*{\argmin}{argmin}

\title{Scaling Submodular Maximization via Pruned Submodularity Graphs}

\vspace{-4mm}
\author{
Tianyi Zhou$^\dagger$, Hua Ouyang$^\ddagger$, Yi Chang$^\ddagger$, Jeff Bilmes$^\S$, Carlos Guestrin$^\dagger$\\
$^\dagger$Computer Science \& Engineering, $^\S$Electrical Engineering, University of Washington, Seattle\\
$^\ddagger$Yahoo Research, Sunnyvale\\
\texttt{\{tianyizh, bilmes, guestrin\}@uw.edu}, \texttt{\{houyang, yichang\}@yahoo\_inc.com}\\
}

\usepackage{ifthen}
\usepackage[usenames,dvipsnames,dvinames]{xcolor}
\providecommand{\dodraft}{true}
\newboolean{isdraft}
\setboolean{isdraft}{\dodraft} 
\ifthenelse{\boolean{isdraft}}{%
\newcommand{\tianyi}[2]{{\color{blue}{#1 $\to$ {\bf Tianyi}}: #2}}
\newcommand{\jeff}[2]{{\color{red}{#1 $\to$ {\bf Jeff}}: #2}}
\newcommand{\todo}[1]{{\color{red}{{\bf TODO: #1} }}}
\newcommand{\fixme}[1]{{\color{red}{{\bf FIXME: #1} }}}

} {

\newcommand{\tianyi}[2]{}
\newcommand{\jeff}[2]{}
\newcommand{\todo}[1]{}
\newcommand{\fixme}[1]{}

}
\makeatletter
\g@addto@macro\normalsize{%
  \setlength\abovedisplayskip{.3ex}
  \setlength\belowdisplayskip{.3ex}
  \setlength\abovedisplayshortskip{.3ex}
  \setlength\belowdisplayshortskip{.3ex}
}
\makeatother

\vspace{-4mm}
\begin{document}
\maketitle

\vspace{-4mm}
\begin{abstract}
  We propose a new random pruning method (called ``submodular
  sparsification (SS)'') to reduce the cost of submodular
  maximization. The pruning is applied via a ``submodularity graph''
  over the $n$ ground elements, where each directed edge is associated
  with a pairwise dependency defined by the submodular function. In
  each step, SS prunes a $1-1/\sqrt{c}$ (for $c>1$) fraction of the
  nodes using weights on edges computed based on only a small number
  ($O(\log n)$) of randomly sampled nodes. The algorithm requires
  $\log_{\sqrt{c}}n$ steps with a small and highly parallelizable
  per-step computation. An accuracy-speed tradeoff parameter $c$, set as $c = 8$, leads
  to a fast shrink rate $\sqrt{2}/4$ and small iteration complexity
  $\log_{2\sqrt{2}}n$. Analysis shows that w.h.p., the greedy
  algorithm on the pruned set of size $O(\log^2 n)$ can achieve a
  guarantee similar to that of processing the original dataset. In
  news and video summarization tasks, SS is able to substantially
  reduce both computational costs and memory usage, while maintaining
  (or even slightly exceeding) the quality of the original (and much
  more costly) greedy algorithm.\looseness-1

%\tianyi{Jeff}{TO ADD: 
%For example, in each step, with a tradeoff parameter set as $c=8$,
%SS prunes a $1-\sqrt{2}/4 \approx 64.6\%$ fraction
%of the nodes using weights on edges computed based on only a small number
%($O(\log n)$) of randomly sampled nodes, and has a small quality
%loss w.h.p.}

\end{abstract}

\vspace{-2mm}
\section{Introduction}
\label{sec:introduction}

Machine learning applications benefit from the existence of large
volumes of data.  The recent explosive growth of data, however, poses
serious challenges both to humans and machines.  One of the primary
goals of a summarization process is to select a representative subset
that reduces redundancy but preserves fidelity to the original
data~\cite{LinSummarization}. Any further processing on only a summary (a small
representative set) by either a human or machine thus reduces
computation, memory requirements, and overall effort.
Summarization has many applications such as news digesting, photo
stream presenting, data subset selection, and video thumbnailing. A summarization algorithm,
however, involves challenging combinatorial optimization problems, whose
quality and speed heavily depend on the objective that assigns quality
scores to candidate summaries.

Submodular functions~\cite{Fujishige,LinSummarization} are broadly applied as objectives for summarization, since they naturally capture
redundancy amongst groups of data elements. A submodular function is a 
set function $f:2^V\rightarrow \mathbb R$ with a diminishing returns property, i.e., 
given a finite ``ground'' set $V$, and any $A\subseteq B\subseteq V$ 
and a $v \notin B$, we have:
\begin{equation}
f(v\cup A) - f(A)\geq f(v\cup B) - f(B).
\end{equation}
This implies $v$ is more important to the smaller set $A$ than to the larger set $B$. The increase $f(v\cup A) - f(A)$ reflects the importance of $v$ to $A$ and is called the ``marginal gain'' $f(v|A)$ of $v$ conditioned on $A$. The objective $f(\cdot)$ can be chosen from a large family of functions (e.g., including but not limited to facility location and set cover functions). 
Usually one requires a small
summary, so a cardinality-based budget is used. Hence, a summarization task
can be cast as the following:
\begin{equation}\label{equ:card}
\max_{S\subseteq V,\\|S|\leq k}f(S).
\end{equation}
Knapsacks and matroids are also often used as constraints. 
In this paper, however, we will primarily be concerned with cardinality constraints, but our methods do generalize to other constraints as well.

Though submodular maximization is NP-hard, a near optimal solution of~\eqref{equ:card} can be achieved via the greedy algorithm, having an approximation factor of $1-1/e$ \cite{greedyapprox}. The greedy algorithm starts with $S\leftarrow\emptyset$, and selects the next element with the largest marginal gain $f(v|S)$ from $V\backslash S$, i.e., $S\leftarrow S\cup \{ v^* \}$ where
$v^* \in \argmax_{v\in V\backslash S}f(v|S)$, and this repeats until $|S|=k$. It is simple to implement and usually outperforms other methods, e.g., those based on integer linear programming.

Scaling up the greedy algorithm to very large data sizes (where $|V|=n$ is big) is a nontrivial practical problem. The per-step computation of greedy is expensive: each step needs to re-evaluate the marginal gains of all elements in $V\backslash S$ conditioned on the new $S$, and thus requires $O(n)$ function evaluations. In addition, each step depends on the results from previous steps, so the computation does not trivially parallelize. Moreover, one typically must keep all $n$ elements in memory until the end of the algorithm, since any element might become the one with the largest marginal gain $f(v|S)$ as $S$ grows.  To overcome this problem, it would be helpful to have an economical screening method to reduce the data size before the costly submodular maximization is performed. While related work is described in \S\ref{sec:related-work}, we next
describe the contributions of this work.

\vspace{-3mm}
\subsection{Main Contribution}
\label{sec:main-contribution}

% Submodular function $f$ as a set function can describe complicated relationship among multiple elements via $f(v|S)$ for element $v\in V$ and subset $S\subseteq V$. Directly maximizing $f$ usually requires precise evaluation of $f(v|S)$ for different $v\in V$ and $S\subseteq V$, which is expensive. However, 
\vspace{-2mm}
A submodular function $f$ can describe higher order relationships among multiple ($\geq3$) elements via $f(v|S)$. In the greedy algorithm, selecting important elements (for maximizing $f$) requires evaluating $f(v|S)$ for all $v\in V\backslash S$ each step. In this paper, we show that removing unimportant elements from $V$ need only use a rough estimate of $f(v|S)$, one that can be derived solely from pairwise relationships $f(v|u)$ for a small set of element pairs $(u,v)$. We encode the pairwise relationships as edge weights on a ``submodularity graph''. By taking advantage of the properties of this graph, the size of the ground set $V$ can efficiently be reduced from $n$ to $O(\log^2n)$ by randomly pruning the nodes on the graph according to a subset of the edge weights.

In particular, given objective $f$, we define a directed submodularity graph whose nodes are the $n$ elements in $V$, and each edge $u \to v$ from tail $u$ to head $v$ is associated with a weight $w_{u \to v} \equiv w_{uv}=f(v|u)-f(u|V\backslash u)$ that reflects the worst-case net loss when maximizing $f$ caused by removing $v$ while retaining $u$ ($f(v|u)$ is the greatest loss when removing $v$ while retaining $u$ while $f(u| V \setminus u)$ is the least gain of retaining $u$).  Intuitively, removing head nodes from $V$ with small-weight edges reduces the ground set from $V$ to a (hopefully much) smaller $V'$, and selecting elements from $V'$ rather than $V$ causes a small overall objective loss but can be much faster.

Finding, however, the smallest $V'\subseteq V$ such that the resulting objective loss can be upper bounded by some constant turns out to be another challenging non-monotone submodular maximization problem, leading to a chicken-and-egg situation. In addition, finding a near optimal solution to this problem requires computing weights on all $n(n-1) = O(n^2)$ edges. We instead propose a randomized pruning method called ``submodular sparsification (SS)'' to reduce the ground set. By leveraging a directed triangle inequality on the submodularity graph (Lemma~\ref{lemma:triangle}), SS only needs to compute partial weights on a few randomly selected edges, and this only slightly increases the objective loss caused by using the reduced set $V'$ rather than $V$. At each step, SS randomly samples $O(\log n)$ elements from $V$ as probes, and removes a $1-1/\sqrt{c}$ fraction of head elements in $V$ that have the smallest weights from amongst the randomly selected elements. When tradeoff parameter $c>1$ increases, the success probability of the randomized algorithm increases, but memory size $|V'|$ also increases. With it set as $c=8$, the number of iterations $\log_{\sqrt{c}}n=\log_{2\sqrt{2}}n$ is small, and per-iteration complexity is dominated by the computation of the pairwise edge weights, which is small and highly parallelizable. Hence, SS can scaled to large data sizes.

In experiments, we compare SS with the lazy greedy and sieve-streaming algorithm \cite{Badanidiyuru} on real-world news and video summarization datasets. Using the lazy greedy algorithm with an SS-reduced ground set, we achieve quality similar to that on the original ground set, but with computation and memory load greedy reduced and, in fact, comparable to a streaming algorithm whose quality is usually much worse than offline methods.

\subsection{Related Work}
\label{sec:related-work}

A number of methods have been proposed to accelerate the speed of the greedy algorithm.  Most of them, however, aim to reduce or distribute the computation rather than the memory, and rarely do they study how to reduce the ground set $V$. Therefore, their contributions are mostly complementary with SS (i.e., they can be combined with SS to further improve algorithmic scalability).

The lazy, or accelerated, greedy algorithm~\cite{lazygreedy, LeskovecLazyGreedy} reduces the number of function evaluations per step by lazily updating a priority queue of marginal gains over all elements. At each step, the algorithm repeatedly updates $f(v|S)$ of the top element and re-inserts it to a queue until the top element does not change position in the queue --- it then adds this element to the running solution. Due to submodularity, the lazy greedy algorithm has the same output and mathematical guarantee as the original greedy algorithm, but significantly reduces computation in practice, but in the worst case it is as slow (if not slower) than the original greedy algorithm.

Approximate greedy algorithms further reduce the number of function evaluations per step at a cost of a worse approximation factor. In~\cite{fast-multi-stage, ApproxGreedy}, each step only approximates identifying the element with the largest marginal gain $\max_{v\in V\backslash S}f(v|S)$ by finding any element whose marginal gain is larger than a fraction $\beta$ of $\max_{v\in V\backslash S}f(v|S)$ of its upper bound. The ``lazier than lazy greedy'' approach~\cite{lazier-than-lazy} selects the element from a smaller random subset $V'\subseteq V\backslash S$ each step, so only the marginal gains of $v\in V'$ need be computed. A similar algorithm in~\cite{Buchbinder2014} randomly selects an element from a reasonably good subset $V'\subseteq V\backslash S$ per step, and extends to the non-monotone case.

Streaming submodular maximization~\cite{Badanidiyuru, Buchbinder, Chekuri, gomes10budgeted, Bateni} studies how to approximate the greedy algorithm in one pass of data under a limited memory budget (i.e., the algorithm can access only a small number of elements in the stream history at a time). The best known approximation factor and hardness are both $1/2$ \cite{Badanidiyuru, Buchbinder}, worse than the $1-1/e$ of the offline greedy algorithm.\looseness-1

Distributed and parallel greedy algorithms~\cite{dist_greedy, parallel_dgreedy} typically partition the ground set into several not-necessarily disjoint pieces and assigns them to multiple machines, then run greedy on each machine, and finally combine the results. These approaches fall into the framework of composable coresets. The existence of such methods for some important submodular maximization problems is not always possible~\cite{compcoreset}. In~\cite{RandomCoreset}, a $1/3$-randomized composable coreset method is proposed to achieve an expected bound for the combined solution. The major difference of this paper is that we study how to reduce the ground set rather than partition it, by developing a coreset-like algorithm on submodularity graph rather than running greedy algorithm to achieve coreset on each machine. However, by replacing the greedy algorithm on each machine with SS, we can further speed up distributed submodular maximization by speeding up the computation at each parallel node.

Another class of methods~\cite{fast-submodular-semigradient, fast-multi-stage} accelerates the greedy algorithm by maximizing a surrogate function whose evaluation is faster and cheaper than the original objective. The surrogate can be either a tight modular lower bound or a simpler submodular function. It can also be adaptively changed in each step to better approach the original objective. In~\cite{fast-multi-stage}, a simple pruning method is used to reduce $V$ by exploiting $f(v|V\backslash v)$, a lower bound of $f(v|S)$ for $S\subseteq V$. E.g., element $u$ whose singleton gain $f(u)$ is less than the $k^{th}$ largest $f(v|V\backslash v)$ over all $v\in V$ can be safely removed. Besides exploiting the global redundancy of $v$ via $f(v|V\backslash v)$, the weight $w_{uv}$ used in SS further takes the pairwise relationship $f(v|u)$ into account. This can result in further ground set reduction.

%Instead of using modular function to 
%
%A broadly observed phenomenon is that the edge measurements on the graph are much easier and cheaper to collect than node features, such as user profile and item properties. Because people are more likely to comments and share stuffs than to reveal their own information. The main challenge of learning the graph based latent variable models in practice is how to efficiently estimate their parameters from partial edge measurements.
%
%A rich class of submodular functions are graph based and need to build a graph before maximizing them.
%
%Submodularity graph contains various structures representing the Markov properties of all paths on it. Exploring it allows us to extract a structured subset for submodular maximization, i.e., rather than just finding a subset maximizing the objective, more local structures inside the subset can be required. This leads to structured summarization in application. One critical example is storyline/timeline generalization.

\section{Submodularity Graph}
\label{sec:submodularity-graph}

We next introduce the ``submodularity graph,'' a useful and efficient tool to explore the redundancy of ground sets $V$ in a submodular maximization process.
\begin{definition}
The submodularity graph is a weighted directed graph $G(V, E,w)$ defined by a normalized submodular function $f:2^V\rightarrow \mathbb R_+$ where $V$ is the set of nodes corresponding to the ground set, 
and each directed edge $e=(u \to v) = (u, v)\in E$ from $u$ to $v$ has weight defined as:
\begin{equation}\label{equ:wuv}
w_{uv}=f(v|u)-f(u|V\backslash u).
\end{equation} 
\end{definition}
Intuitively, the weight $w_{uv}$ measures the worst case net loss in maximizing $f(S)$ on a reduced set $V'$ with $v$ removed and $u$ retained. In Eq.~\eqref{equ:wuv}, $f(v|u)$ is the maximum possible gain $v$ can offer a set involving $u$, while $f(u|V\backslash u)$ is the minimal possible gain $u$ can contribute to the solution $S$ because $f(u|S)\geq f(u|V\backslash u)$ holds by submodularity. Hence, a small $f(v|u)$ indicates $v$ is unimportant if $u$ is retained in a solution, while a large $f(u|V\backslash u)$ implies that $u$ is always important. Taken together, a small $w_{uv}$ would suggest removing $v$ while keeping $u$. Note $w_{uv}$ is a net loss, combining both the ``local'' importance of $f(v|u)$ and the ``global'' importance of $f(u|V\backslash u)$. Previous work such as~\cite{fast-multi-stage} and curvature based methods~\cite{iyer2013-curvature} do not leverage local and global importance in the same way.

We further generalize $G(V, E)$ to a ``conditional submodularity graph'' $G(V, E|S)$ describing the pairwise relationships conditioned on set $S\subseteq V$. Accordingly, the edge weight on $e=(u, v)$ is:
\begin{equation}\label{equ:wuvS}
w_{uv|S}=f(v|S + u)-f(u|V\backslash u).
\end{equation}
$G(V, E|S)$ reduces to $G(V, E)$ when $S=\emptyset$, usually the starting set
in a greedy submodular maximization procedure. Below we give a detailed analysis of how edge weight $w_{uv}$ can be used to remove elements from $V$. For notational simplicity, we use ``$+$'' to denote the set union ``$\cup$,'' and ``$-$'' for set subtraction ``$\backslash$''. We start by studying two properties of $w_{uv|S}$.
\begin{lemma}\label{lemma:wSwP}
If $P\subseteq S\subseteq V$, for any $u, v\in V$ such that $u, v\notin S$, $w_{uv|S}\leq w_{uv|P}$.
\end{lemma}
\vspace{-3ex}
\begin{proof}
Submodularity requires $f(v|S+u)\leq f(v|P+u)$. From the definition of $w_{uv|S}$ in~\eqref{equ:wuvS}, the conclusion is immediate.
\end{proof}
\begin{lemma}\label{lemma:wuv}
For any $u, v\in V$ and $S\subseteq V$, if $u\neq v$ and $u, v\notin S$, then
\begin{equation}
f(v|S)\leq f(u|S)+w_{uv|S}.
\end{equation}
\end{lemma}
\vspace{-2ex}
\begin{proof}
\begin{align}
\label{equ:price}
f(v|S) &= f(u|S)+f(v|u+S)-f(u|v+S) \\
       &\leq f(u|S)+f(v|u+S)-f(u|V-u)=f(u|S)+w_{uv|S}.
\end{align}
The first equality is obtained using the definition of the marginal gain, while the inequality is from submodularity and since $(v+S)\subseteq (V-u)$.
\vspace{-2mm}
\end{proof}
\vspace{-2mm}
Lemma~\ref{lemma:wuv} states that the weight $w_{uv}$ relates the 
two marginal gains of $u$ and $v$ relative to $S$. The marginal gain $f(v|S)$ plays a critical role in various submodular maximization algorithms since it measures how much $f(S)$ is improved by adding $v$ to $S$. In each step, the greedy algorithm selects the element with the largest $f(v|S)$, i.e., $S\leftarrow \argmax_{x\in V}f(x|S)\cup S$, and $f(S)$ increases by $f(v|S)$.

If $v\in \argmax_{x\in V \setminus S}f(x|S)$ should be selected by the greedy algorithm at the current step, but for some reason is missing in $V' \subseteq V$ (a reduced ground set), then greedy instead selects $u \in \argmax_{x\in V'}f(x|S)$. In this case, the objective $f(S)$ increases by $f(u|S) \leq f(v|S)$ rather than $f(v|S)$. 
By the relative optimality of $u$ in $V'$ and Lemma~\ref{lemma:wuv}, we have
\begin{equation}\label{equ:loss}
f(u|S)\geq f(\argmin\limits_{x\in V'}w_{xv|S}|S)\geq f(v|S)-\min\limits_{x\in V'}w_{xv|S}.
\end{equation}
Hence, the objective loss caused by removing $v$ from $V$ and using $u$ instead is at most
the minimal weight over all edges entering $v$ from other elements in
$V'$. In other words, an upper bound on the price for pruning $v$ is $\min_{x\in
  V'}w_{xv|S}$, which reflects the contribution of $v$ to the set
$V'$. If it is small, the objective loss is, relatively speaking, negligible and $v$ 
may be removed with impunity. We hence define this concept as a
``divergence'' of $v$ from $V'$ on $G(V, E|S)$:
\begin{definition}\label{def:wVv}
On the submodularity graph $G(V, E)$, the divergence $w_{V',v}$ of a node $v\in V$ from a set of nodes $V'$ is defined as $w_{V',v}=\min_{x\in V'}w_{xv}$. Similarly, the divergence $w_{V',v|S}$ on the conditional submodularity graph $G(V, E|S)$ is defined as $w_{V',v|S}=\min_{x\in V'}w_{xv|S}$.
\end{definition}
Although the edge weights $w_{uv}$ are asymmetric, we next show that a directed triangle inequality holds on $G(V, E)$. This plays significant role in SS, since it provides an upper bound on an edge weight based on weights of adjacent edges, and thus avoids needing to compute all the edge weights exactly.\looseness-1
\begin{lemma}\label{lemma:triangle}
For $u,v,x\in V$, we have $w_{vx}\leq w_{vu}+w_{ux}$.
\end{lemma}
\vspace{-1ex}
The proof is given in \cite{Supp}. A similar inequality also holds for 
$w_{uv|S}$ defined on $G(V, E|S)$.

\section{Submodular Sparsification}
\label{sec:subm-spars}

In this section, we introduce submodular sparsification (SS), a
randomized pruning algorithm that reduces $V$ to $V'\subseteq V$
without drastically hurting the optimality of submodular
maximization. Although pruning the conditional submodularity graph
$G(V, E|S)$ with the greedy algorithm can rule out additional
elements, here we focus on reducing $V$ before running any submodular
maximization algorithm, i.e., when $S=\emptyset$, but it is worth noting
that SS can be easily extended to $G(V, E|S)$.

\subsection{Pruning as Submodular Maximization}
\label{sec:prun-as-subm}

According to Eq.~\eqref{equ:loss} and Definition~\ref{def:wVv}, small $w_{V'v}$ for all pruned elements $v\in V\backslash V'$ leads to small loss in the per-step increase of objective function by the greedy algorithm. By parameterizing an upper bound $w_{V'v}\leq \epsilon$, the following seeks the best pruned set $V'$ for use in the maximization of $f$.
\begin{definition}[submodular sparsification]
The submodular sparsification problem is to solve:
\begin{equation}\label{equ:problemss}
\max_{V'\subseteq V} h(V') := \left|\left\{v\in V\backslash V': w_{V'v}\leq\epsilon\right\}\right|.
\end{equation}
\end{definition}
\begin{proposition}\label{prop:sssm}
The objective function $h(\cdot)$ in Eq.~\ref{equ:problemss} is non-monotone submodular.
\end{proposition}
\vspace{-3mm}
The proof is in \cite{Supp}. Let $V^*$ of size $K \triangleq|V^*|$ be the optimal solution of Eq.~\eqref{equ:problemss} (note all are $\epsilon$-dependent, the proof also shows $h$ is monotone in $\epsilon$). 
Running greedy on $V^*$ rather than $V$ yields: 
\begin{theorem}\label{the:ssbound}
Let $S^* \in \argmax_{S\subseteq V, |S|\leq k} f(S)$, where $f: 2^V\rightarrow \mathbb R_+$ is normalized non-decreasing and submodular, let $S'$ be a greedy solution to the problem $\max_{S\subseteq V^*, |S|\leq k} f(S)$. If $|V^*|\geq k$, the following approximation bound holds for $S'$.
\begin{equation}
f(S')\geq \left(1-e^{-1}\right)\left(f(S^*)-k\epsilon\right).
\end{equation}
\end{theorem}
A proof of this is given in \cite{Supp}.
Unfortunately, solving Eq.~\eqref{equ:problemss} leads to a chicken-and-egg problem: even approximately solving this unconstrained non-monotone submodular maximization requires an expensive bi-directional randomized greedy algorithm~\cite{DoubleGreedy} having approximation factor $1/2$ and that is slow in practice. Also, when $f$ is not a graph based submodular function (such as facility location or saturated coverage), solving Eq.~\eqref{equ:problemss} requires a costly computation of the weights on all $n(n-1)$ edges.\looseness-1

\subsection{Randomized Pruning} 
\label{sec:randomized-pruning}

Drawing inspiration from bi-criteria $k$-clustering in Euclidean
space~\cite{BiClustering}, we develop a randomized pruning method
(``submodular sparsification (SS)'') on a submodularity graph to produce a
reduced ground set $V'$ without either computing all $n(n-1)$ weights
or running bi-directional greedy.
\vspace{-2mm}
\begin{algorithm}[htb]
	\caption{Submodular Sparsification (SS)}
	\begin{algorithmic}[1]\label{alg:coreset}
		\STATE \textbf{Input:} $V$, $f$, $r$, $c$
		\STATE \textbf{Output:} $V'$
		\STATE \textbf{Initialize:} $V'\leftarrow\emptyset$, $n\leftarrow|V|$
		\WHILE{$|V|>r\log n$}
		\STATE Sample $r\log n$ items uniformly at random from $V$ and place them in $U$;
		\STATE $V\leftarrow V\backslash U$;
		\STATE $V'\leftarrow V' \cup U$;
%		\FOR {$v\in V$, $u\in R$}
%		\IF{$f(v|u+S)\leq\epsilon$}\STATE{$V:=V\backslash v$}\ENDIF
%		\ENDFOR
		\FOR {$v\in V$}\STATE{$w_{U,v}\leftarrow\min\limits_{u\in U}[f(v|u)-f(u|V\backslash u)]$}\ENDFOR
		\STATE Remove from $V$ the top $(1-1/\sqrt{c})|V|$ of elements with the smallest $w_{Uv}$;
		\ENDWHILE
		\STATE $V'\leftarrow V\cup V'$
	\end{algorithmic}
\end{algorithm}
The submodular sparsification procedure is given in Algorithm~\ref{alg:coreset}. It starts from the original ground set $V$ and an empty set $V'$. At each iteration, it randomly samples a size-$(r\log n)$ set\footnote{The base of all logarithms in this paper is $2$ if not otherwise specified.} of elements $U$ from the current $V$, acting as probes to test the redundancy of the remaining elements in $V$, that are removed from $V$ and added to $V'$. It then removes the top $(1-1/\sqrt{c})|V|$ elements from $V$ having the smallest divergence $w_{Uv}$ from $U$ on $G(V, E)$ because of their unimportance to $U$. The procedure repeats and the size of $V$ shrinks exponentially fast (with a shrink rate of $1/\sqrt{c}$) until it falls below a threshold.
The parameter $r$ controls the size of a probe set $U$ and influences the size of the final $V'$. In our analysis below, $r=O(cK)$ for $c>1$ to produce a sufficiently large success probability. In practice, we choose $c=8$ to produce a fast shrink rate $1/\sqrt{c}=\sqrt{2}/4<1/2$, since it can remove more than half ($\approx 64.6\%$) of $V$ per step. With $r=O(cK)$, since $K$ is unknown in practice, we find that $r=8$, also, empirically works well (see Section~\ref{sec:experiments}).
% I guess 8 is the lucky number!

Algorithm~\ref{alg:coreset} finishes in $\log_{\sqrt{c}}n$ iterations. It leads to small iteration complexity $\log_{2\sqrt{2}}n$ when $c=8$. The per iteration computation is dominated by computing $w_{U,v}$, which requires calculating $O(n\log n)$ pairwise relationships. This can be simplified if $f$ is graph based, because the first $O(n)$ greedy step already requires all of the pairwise similarities/distances needed for further $f$ evaluations. When $f$ is not graph based, this can be accelerated via parallelization, since disjoint pairs $u,v$ in the set $\{ f(u|v) \}_{u,v}$ may be independently computed. $f(u|V\setminus u)$ may be precomputed once in linear time.\looseness-1

\subsection{Analysis of Submodular Sparsification}
\label{sec:analys-subm-spars}

According to Lemma~\ref{lemma:wuv}, a small $w_{uv}$ leads to a small objective loss when $v$ is removed and $u$ retained. 
Instead of solving non-monotone submodular maximization in
Eq.~\eqref{equ:problemss}, SS randomly selects probes $u \in U$ to rule out elements $v$ from $V$. The following lemma uses the directed triangle inequality in Lemma~\ref{lemma:triangle} to study which $u$'s, if sampled, can lead to a relatively small $w_{uv}$ and thus a small $w_{Uv}$ in Algorithm~\ref{alg:coreset}. Proofs of all the following results can be found in \cite{Supp}.
\begin{lemma}\label{lemma:uv}
Let $u^*_v \in \argmin_{u\in V^*}w_{uv}$ be the tail node of an edge with the minimal weight over all edges from elements in $V^*$ to head $v$.
Then, for any item $v$, $\forall u\in P(u^*_v)\cap Q(u^*_v)$ where
	\begin{align}
	\notag &P(u^*_v)=\{u\in V:f(u+u^*_v)\leq f(v+u^*_v)\},\\
	\notag &Q(u^*_v)=\{u\in V:f(u)+f(u|V\backslash u)\geq f(u^*_v)+f(u^*_v|V\backslash u^*_v)\}.
	\end{align}
	we have that $w_{uv}\leq 2w_{u^*_vv}$.
\end{lemma}
Lemma~\ref{lemma:uv} states that for any item $v$, if $P(u^*_v)\cap Q(u^*_v)\neq\emptyset$ and at least one $u\in P(u^*_v)\cap Q(u^*_v)$ is sampled in Algorithm~\ref{alg:coreset}, then $w_{uv}$, the maximal loss in $f(S)$ caused by dropping $v$, is sufficiently small, so $v$ can be safely removed. The below discusses how to sample $u$s and drop $v$s.
\begin{proposition}\label{prop:outsideball}
	For an element $u^*\in V^*$ and $c>1$, define its $|V|/(cK)$-NN ball $B\left(u^*,|V|/(cK)\right)$ as the set of $|V|/(cK)$ elements in $V$ with the smallest $f(u+u^*)$, and let $V_{u^*}=\{v\in V:u^*_v=u^*\}$ denote the set of elements ruled out by $u^*$. If one $u\in B(u^*,|V|/(cK))\cap Q(u^*)$ is sampled into $U$ in some iteration of Algorithm~\ref{alg:coreset}, then all the elements in $V_{u^*}$ outside the ball fulfill the following:
\abovedisplayshortskip=.2ex\belowdisplayshortskip=.2ex\abovedisplayskip=.2ex\belowdisplayskip=.2ex%
\begin{align}
	\forall v\in V_{u^*}\backslash B\Bigl(u^*,|V|/(cK)\Bigr), \;\; w_{uv}\leq 2w_{u^*v}.
      \end{align}
\end{proposition}\vspace{-2ex}
Based on Proposition~\ref{prop:outsideball}, we can derive the maximal number of removed elements $v$ whose importance represented by $w_{Uv}$ cannot be upper bounded.
\begin{proposition}\label{prop:upper4bad}
	For each $u^*\in V^*$, if one $u\in B\left(u^*,|V|/(cK)\right)\cap Q(u^*)$ is sampled into $U$ and added to $V'$ in some iteration of Algorithm~\ref{alg:coreset}, then
	\begin{equation}
	\left|\{x\in V:w_{Ux}\geq 2w_{V^*x}\}\right|\leq |V|/(cK).
	\end{equation}
\end{proposition}\vspace{-2ex}
The following proposition explains why Algorithm~\ref{alg:coreset} reduces ground set $V$ exponentially by a ratio of $1-1/\sqrt{c}$. It also shows that all the pruned elements $v$ satisfy $w_{Uv}\leq 2w_{V^*v}$, which indicates that ruling out them from $V$ will lead to at most a $2w_{V^*v}$ loss in objective $f(S)$.
\begin{proposition}\label{prop:number_v}
	Before line 11 of Algorithm~\ref{alg:coreset}, the following holds.
	\begin{equation}
	\left|\{v\in V:w_{Uv}\leq 2w_{V^*v}\}\right|\geq \left(1-1/\sqrt{c}\right)|V|.
	\end{equation}
\end{proposition}\vspace{-2ex}
Therefore, it is safe to remove the $1-1/\sqrt{c}$ fraction of items from $V$ with the smallest $w_{Uv}$, since their importance $w_{Uv}$ can be upper bounded. 
Proposition~\ref{prop:number_v} results in the following Lemma.
\begin{lemma}\label{lemma:condition}
	For each $u^*\in V^*$, if at least one $u\in B\left(u^*,|V|/(cK)\right)\cap Q(u^*)$ is sampled and added into $U$, $\forall v\in V\backslash V'$ where $V'$ is the output of Algorithm~\ref{alg:coreset}, we have $w_{V'v}\leq 2w_{V^*v}$.
\end{lemma}

\begin{wrapfigure}[13]{r}{0.43\linewidth}
\vspace{-8mm}
\begin{center}
% \vspace{-6mm}
 \includegraphics[width=1\linewidth]{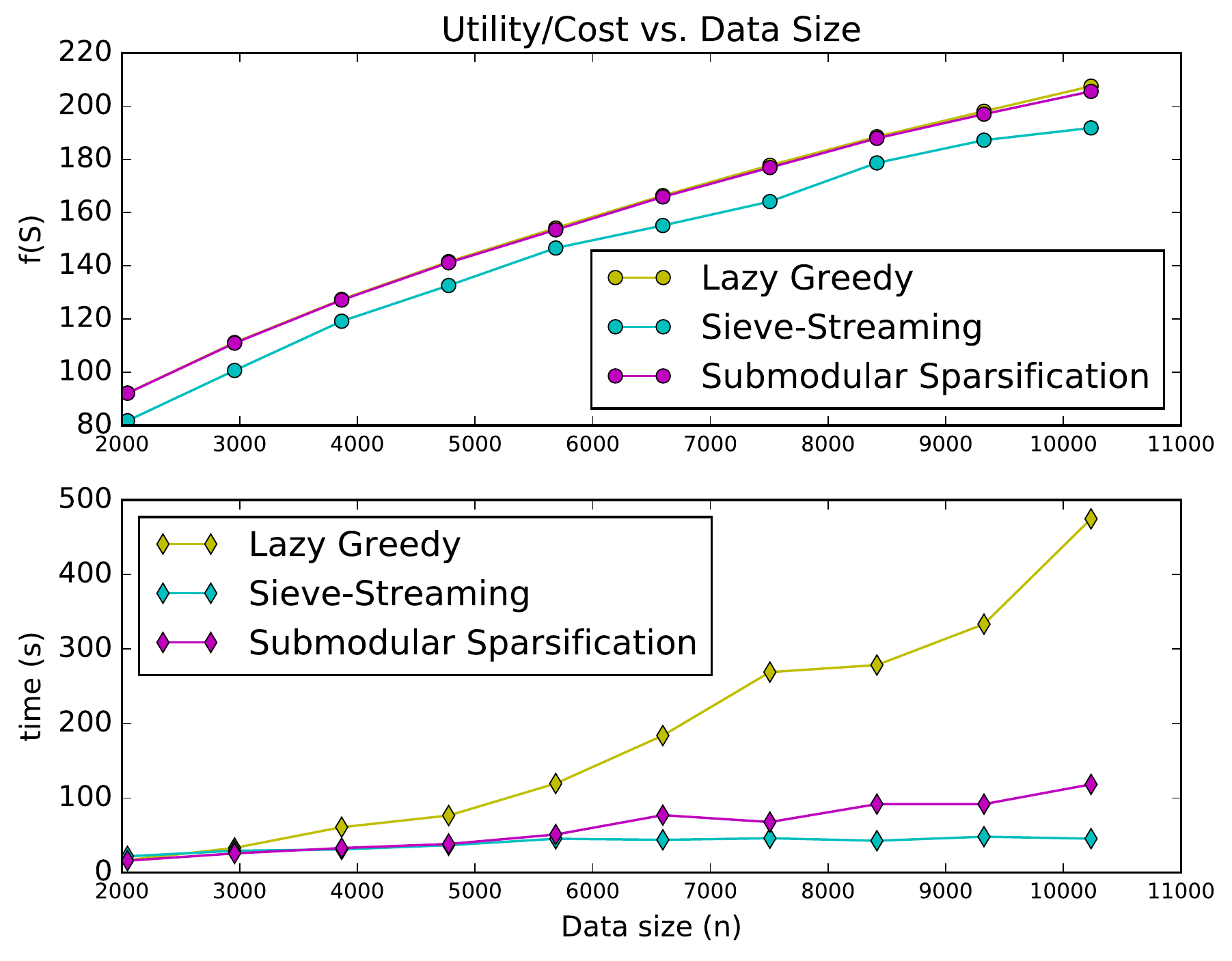}
\end{center}\vspace{-5mm}
   \caption{\scriptsize{Utility $f(S)$ and time cost vs. size of data $n$}}
 \label{fig:nchange}
\vspace{-8mm}
\end{wrapfigure}
Now we study the failure probability, i.e., the probability that the condition in 
Lemma~\ref{lemma:condition} is not true.
\begin{proposition}\label{prop:upperpr}
	If for each $u^*\in V^*$, the probability that sampling an item $u$ uniformly from $V$ such that $f(u)+f(u|V\backslash u)>f(u^*)+f(u^*|V\backslash u^*)$ is not less than $q$, and if  $r=O(cK)=pcK$, then the probability that no $u\in B\left(u^*,|V|/(cK)\right)\cap Q(u^*)$ is sampled and added into $U$ for at least one $u^*\in V^*$ in at least one iteration of Algorithm~\ref{alg:coreset} is at most $n^{1-qp}\log_{\sqrt{c}}n$.
\end{proposition}\vspace{-2ex}

% The assumption in Proposition \ref{prop:upperpr} rules out the case when there are very few elements $u$ whose singleton gain $f(u)\geq f(u^*), \forall u^*\in V^*$. The assumption is reasonable because if this is the case, it is trivial to gain an optimal $f(S)=f(S^*)$ by simply picking up the top $k$ elements with the largest singleton gain $f(u)$, and any pruning is not necessary.
%Here $q$ is the minimal probability that a uniform sample from $V$ is in $Q(u^*)$ over all $u^*\in V^*$.

By using Lemma~\ref{lemma:condition} and Proposition~\ref{prop:upperpr}, we replace $\epsilon$ in the proof of Theorem~\ref{the:ssbound} with $2\epsilon$, which yields:
\begin{theorem}\label{the:app}
	Under the assumptions in Proposition~\ref{prop:upperpr}, the size of the output $V'$ of Algorithm~\ref{alg:coreset} is $|V'|=(cp/\log\sqrt{c})K\log^2n$. With high probability, i.e., $1-n^{1-q p}\log_{\sqrt{c}}n$, we have that $\forall v\in V\backslash V'$, $w_{V'v}\leq 2w_{V^*v}$, and thus the greedy algorithm on $V'$ outputs a solution $S'$ such that
	\begin{equation}\label{equ:probbound}
	f(S')\geq \left(1-e^{-1}\right)\left(f(S^*)-2k\epsilon\right),
	\end{equation}
	where $S^*$ is the optimal solution to Eq.~\eqref{equ:card}, and $k$ is the budget in Eq.~\eqref{equ:card}.
\end{theorem}

\textbf{Remarks:} Critically, via $\epsilon$ and $c$, the above analysis shows 
a tradeoff between: 1) the approximation bound, 2) the size of $V'$ (the memory load), and 3) the computational cost. The approximation bound Eq.~\eqref{equ:probbound} can be improved if $\epsilon$ in Eq.~\eqref{equ:problemss} is small, but a smaller $\epsilon$ leads to larger $K=|V^*|$ (size of the optimal solution to Eq.~\eqref{equ:problemss}). This results in a larger reduced set $V'$ of size $(cq/\log\sqrt{c})K\log^2n$; and a larger $V'$ produced by Algorithm~\ref{alg:coreset} means more computation per step. It also shows a tradeoff between the success probability and $|V'|$ (the memory) via $c$: if $c$ is large, the success probability $1-n^{1-q p}\log_{\sqrt{c}}n$ increases, but $|V'|$ also increases. Note that $\epsilon$ measures the loss from approximate optimality (the $1-1/e$ guarantee), and $K\in[1,|V|]$ measures the $\epsilon$-reducibility of $V$. SS fails when $K=|V|$. On real datasets we observe $|V'| \ll |V|$ even when $\epsilon$ is small, thus suggesting a large zone of practical success for SS.

\begin{wrapfigure}[13]{r}{0.43\linewidth}
\vspace{-2ex}
\begin{center}\vspace{-7mm}
 \includegraphics[width=1\linewidth]{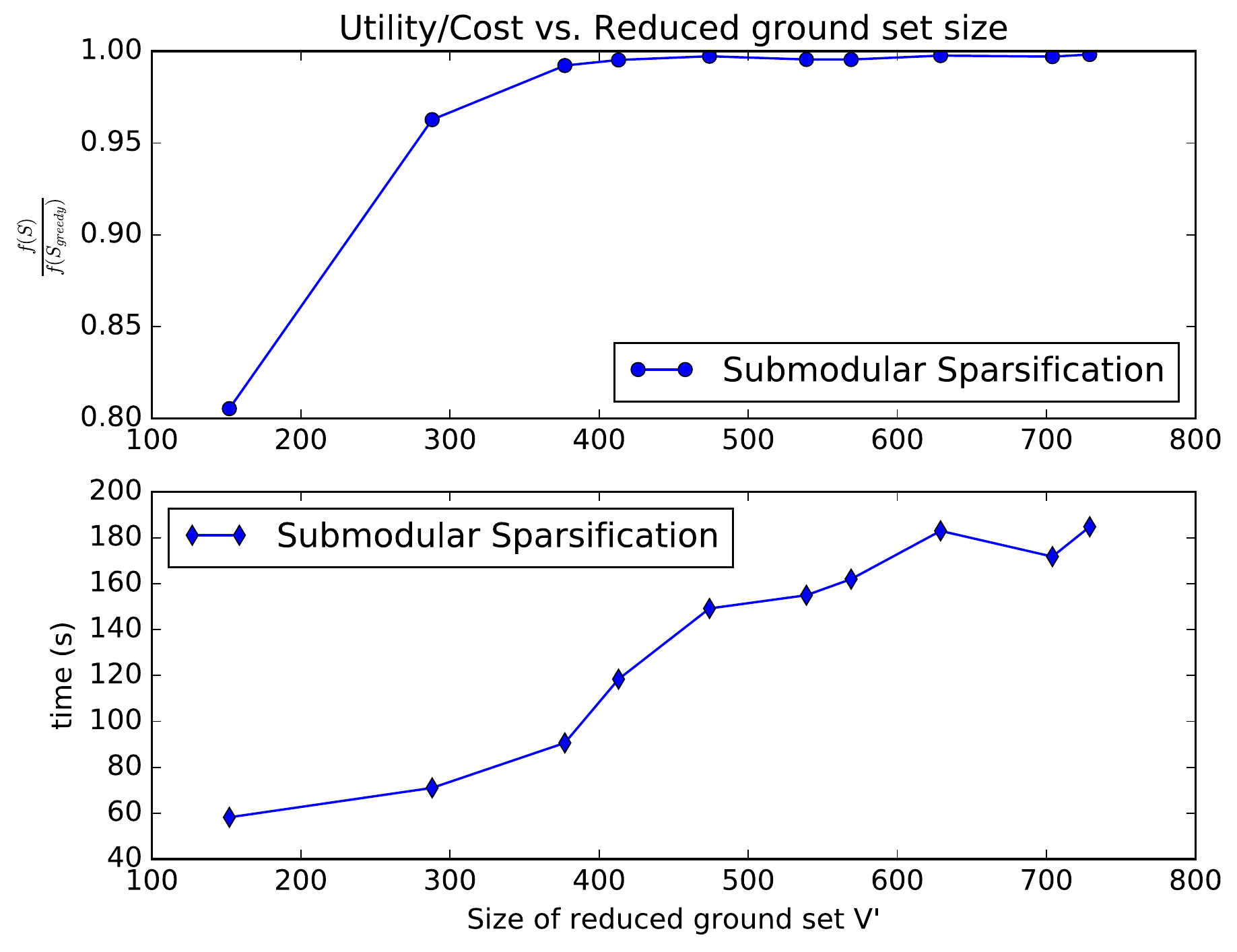}
\end{center}\vspace{-3mm}
   \caption{\scriptsize{Relative utility $f(S)/f(S_{greedy})$ and time cost associated with different sizes of reduced set $V'$, which correspond to $10$ different values of $r$ varying between $[2, 20]$ with step size $2$.}}
 \label{fig:mchange}
\vspace{-3mm}
\end{wrapfigure}
SS can also reduce the ground set for non-monotone submodular maximization monotone under general constraints (e.g., knapsack or matroid) by applying it before any algorithm runs. All previous analysis still holds in general except 
Theorem~\ref{the:ssbound} and Theorem~\ref{the:app}, whose proofs rely on a cardinality constraint and monotonicity. They can be easily modified, however, by applying Eq.~\eqref{equ:viui} to the proof the other algorithm's bound. The fundamental reason is that the properties (Lemmas~\ref{lemma:wSwP}-\ref{lemma:triangle}) of weight $w_{uv}$ on the submodularity graph $G(V, E)$ depend {\em only} on submodularity and non-negativity of $f$.

\subsection{Additional Improvements}
\label{sec:addit-impr}

In practice, several techniques can be further applied to Algorithm~\ref{alg:coreset} to improve either its effectiveness or efficiency. Firstly, the pruning technique based on $f(u|V\backslash u)$ proposed in \cite{fast-multi-stage} can be applied to $V$ before running Algorithm~\ref{alg:coreset} to rule out additional elements and save computation.

\begin{wrapfigure}[12]{r}{0.4\linewidth}
\vspace{-2ex}
\begin{center}\vspace{-5mm}
 \includegraphics[width=1\linewidth]{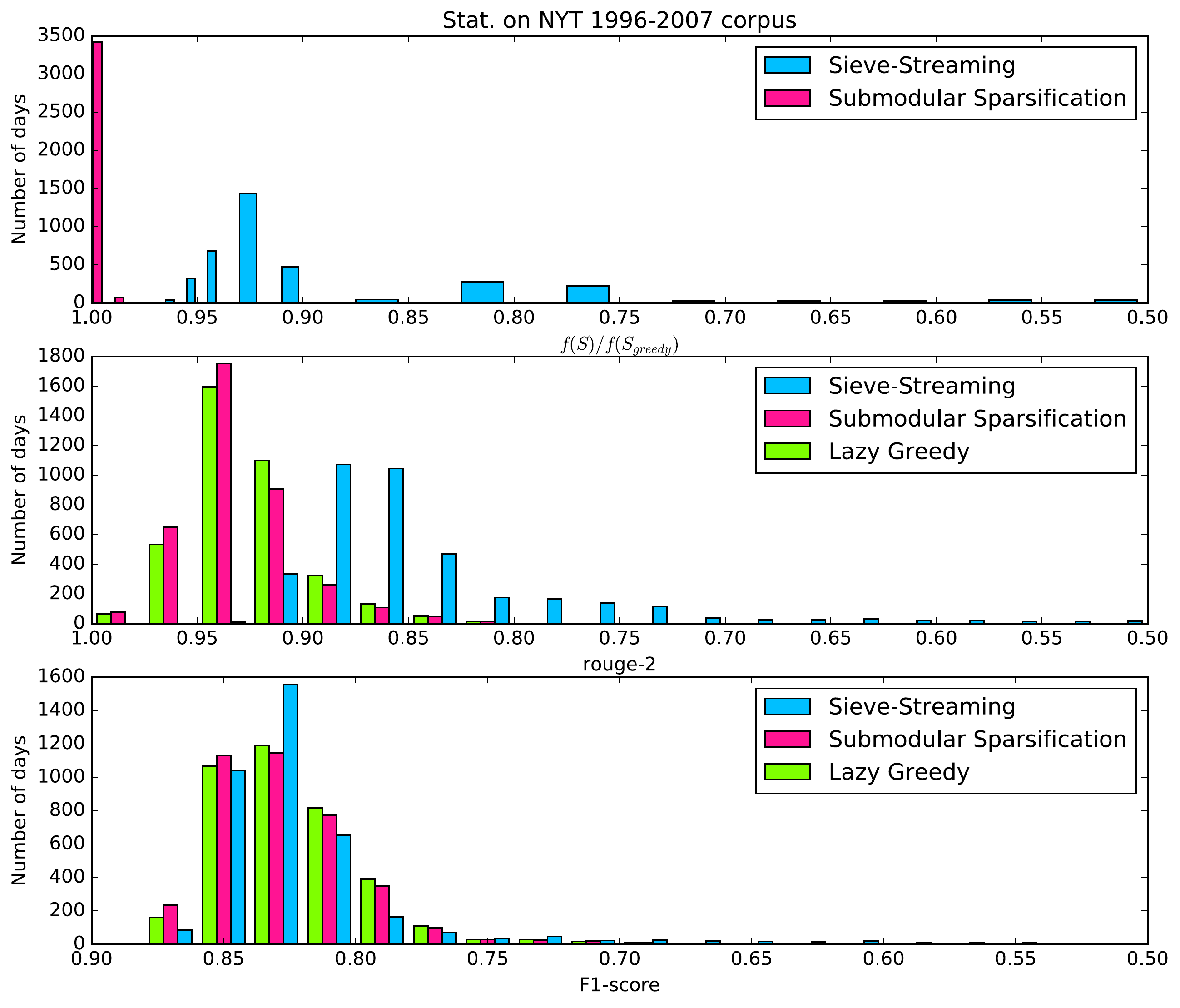}
\end{center}\vspace{-5mm}
   \caption{\scriptsize{Statistics of relative utility $f(S)/f(S_{greedy})$, ROUGE-2 score and F1-score on daily news summarization results of $3823$ days' news from New York Times corpus between 1996-2007.}}
 \label{fig:NYT_utility}
\end{wrapfigure}
The second improvement would use importance rather than uniform sampling in 
Algorithm~\ref{alg:coreset}. According to Proposition~\ref{prop:upperpr}, sampling $u$ with large $f(u)+f(u|V\backslash u)$ is helpful to increase the probability of $u\in Q(u^*_v)$ and $q$, which leads to a larger success probability $1-n^{2-q c}$. Intuitively, large $f(u)$ suggests $u$ may be important, while large $f(u|V\backslash u)$ indicates its importance is undiminished by other elements in $V$.

The third strategy is to further reduce $V'$ by exploring its redundancy. In particular, after Algorithm~\ref{alg:coreset}, the bi-directional greedy algorithm \cite{DoubleGreedy} can be used to solve Eq.~\eqref{equ:problemss} defined on the reduced ground set $V'$. Since $V'$ is much smaller than $V$, the cost may be acceptable.

\vspace{-.5\baselineskip}
\section{Experiments}
\label{sec:experiments}
\vspace{-.5\baselineskip}

%We study how the utility (i.e., $f(S)$ on output $S$) and time cost of different algorithms vary with data size $n$ and the memory size $m$ (for SS $m$ is the size of $V'$). 
In this section, on several news and video datasets, we compare the summary achieved by running the greedy algorithm on the reduced set $V'$ of SS with summaries achieved by other algorithms on the original set $V$. We use the feature based submodular function $f(S)=\sum_{u\in \mathcal U}\sqrt{c_u(S)}$ as our objective, where $\mathcal U$ is a set of features, and $c_u(S)=\sum_{v\in S} \omega_{v, u}$ is a modular score ($\omega_{v, u}$ is the affinity of element $v$ to feature $u$). This function typically achieves good performance on summarization tasks. Our baseline algorithms are the lazy greedy approach~\cite{lazygreedy} (which has identical output as greedy but is faster) and the ``sieve-streaming''~\cite{Badanidiyuru} approach for streaming submodular maximization, which has low memory requirements as it takes one pass over the data. We set $r=8$ and $1-1/sqrt{c} 
= 1-\sqrt{2}/4 \approx 64.6\%$ in Algorithm \ref{alg:coreset}.

%\begin{figure}[htp]\vspace{-3mm}
% \begin{center}
%  \includegraphics[width=1\linewidth]{nchange.pdf}
%%  \includegraphics[width=1\linewidth]{nchange.png}
%%\vspace{.5in}
% \end{center}\vspace{-3mm}
%   \caption{Utility $f(S)$ and time cost vs. size of data $n$}
% \label{fig:nchange}
%\end{figure}\vspace{-3mm}

\subsection{Empirical Study on News}
\label{sec:empirical-study}

%\begin{figure}[h]\vspace{-3mm}
% \begin{center}
%  \includegraphics[width=1\linewidth]{mchange.pdf}
%%  \includegraphics[width=1\linewidth]{mchange.png}
%%\vspace{.5in}
% \end{center}\vspace{-5mm}
%   \caption{Relative utility $f(S)/f(S_{greedy})$ and time cost associated with different sizes of reduced set $V'$, which correspond to $10$ different values of $r$ varying between $[2, 20]$ with step size $2$.}
% \label{fig:mchange}
%\end{figure}\vspace{-3mm}

%\begin{figure}[h]\vspace{-2mm}
% \begin{center}
%  \includegraphics[width=1\linewidth]{NYT_utility.pdf}
%%  \includegraphics[width=0.75\linewidth]{NYT_utility.png}
%%\vspace{.5in}
% \end{center}\vspace{-5mm}
%   \caption{Statistics of relative utility $f(S)/f(S_{greedy})$, ROUGE-2 score and F1-score on daily news summarization results of $3823$ days' news from New York Times corpus between 1996-2007.}
% \label{fig:NYT_utility}
%\end{figure}\vspace{-2mm}

Figure~\ref{fig:nchange} shows how $f(S)$ and time cost varies when we change $n$. The budget size $k$ of the summary set to the number of sentences in a human generated summary. The number of trials in sieve-streaming is $50$, leading to memory requirement of $50k$. The utility curve of SS overlaps that of lazy greedy, while its time cost is much less and increases more slowly than that of lazy greedy. Sieve-streaming performs much worse than SS in terms of utility, and its time cost is only slightly less (this is because it quickly fills $S$ with $k$ elements and stops much earlier before seeing all $n$ elements).
Figure~\ref{fig:mchange} shows how relative utility $f(S)/f(S_{greedy})$ ($S_{greedy}$ is the greedy solution) and SS time cost vary with the size of the reduced set $V'$. SS quickly reaches a $f(S)=0.97f(S_{greedy})$ once the size exceeds $300$, while its computational cost increases slowly. 

% Hence, it requires a few extra computations, and can largely reduces the ground set $V$ and the % memory load. 

%\begin{figure}[htp]\vspace{-3mm}
% \begin{center}
%  \includegraphics[width=0.93\linewidth]{NYT_nvstime.pdf}
%%  \includegraphics[width=1\linewidth]{NYT_nvstime.png}
%%\vspace{.5in}
% \end{center}\vspace{-5mm}
%   \caption{Size of data $n$ vs. time cost on daily news summarization results of $3823$ days' news from New York Times corpus between 1996-2007. The area of each circle is proportional to the relative utility $f(S)/f(S_{greedy})$.}
% \label{fig:NYT_nvstime}
%\end{figure}\vspace{-4mm}

%\begin{figure}[h]\vspace{-2mm}
% \begin{center}
%  \includegraphics[width=0.93\linewidth]{NYT_mvn.png}
%%\vspace{.5in}
% \end{center}\vspace{-5mm}
%   \caption{Scatter plot of relative utility $f(S)/f(S_{greedy})$ achieved by submodular sparsification on the $3823$ days' news with the corresponding size of ground set $V$ and the size of reduced set $V'$. Each point corresponds to one day.}
% \label{fig:NYT_mvn}
%\end{figure}\vspace{-2mm} 

\vspace{-0.5\baselineskip}
\subsection{News Summarization}
\label{sec:news-summarization}

\begin{wrapfigure}[13]{r}{0.43\linewidth}
\vspace{-4em}
\begin{center}\vspace{-4mm}
 \includegraphics[width=1\linewidth]{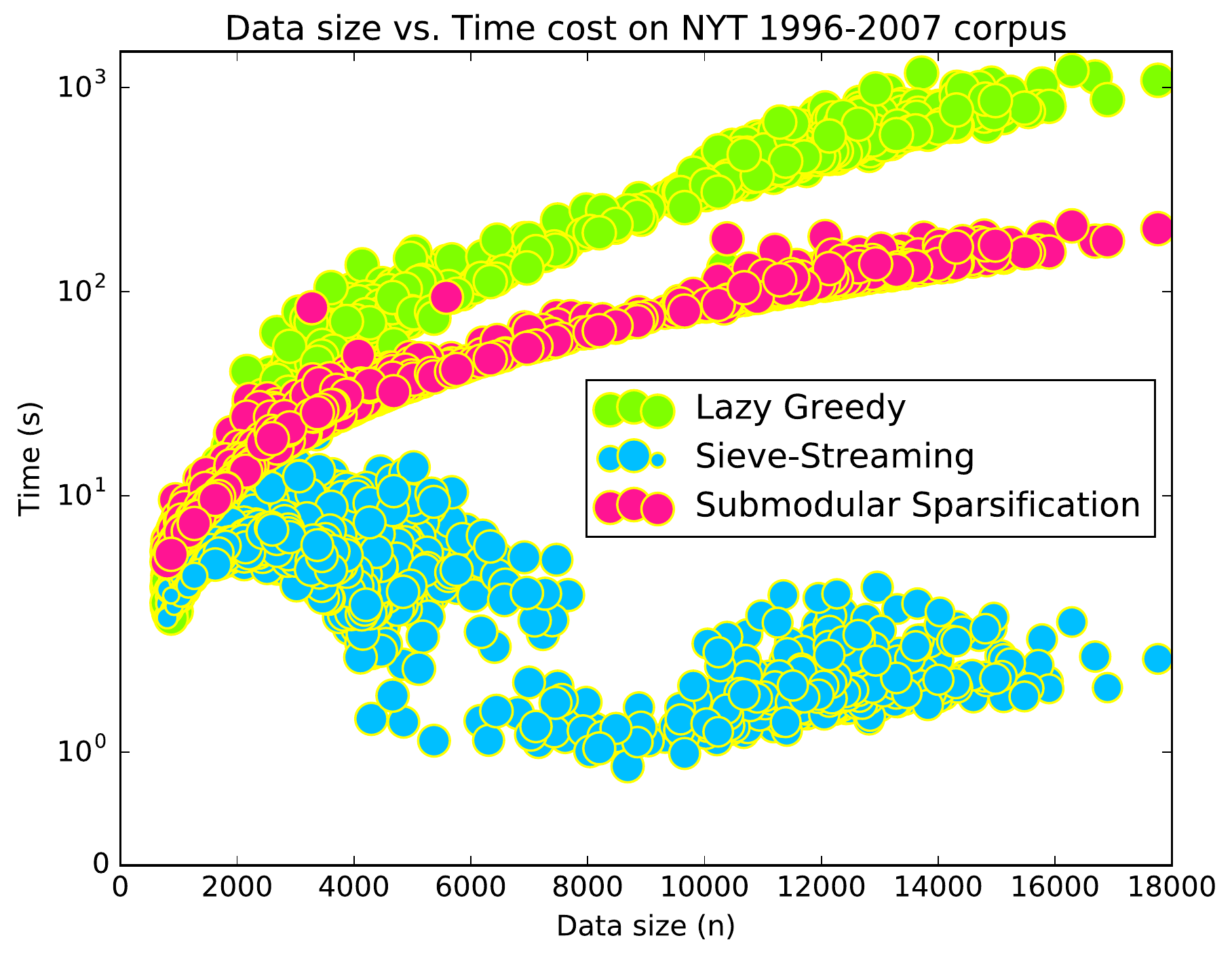}
\end{center}\vspace{-3mm}
   \caption{\scriptsize{Size of data $n$ vs. time cost on daily news summarization results of $3823$ days' news from New York Times corpus between 1996-2007. The area of each circle is proportional to the relative utility $f(S)/f(S_{greedy})$.}}
 \label{fig:NYT_nvstime}
\vspace{-6mm}
\end{wrapfigure}
We conduct summarization experiments on two large news corpora, The
NYTs annotated corpus 1996-2007
(\url{https://catalog.ldc.upenn.edu/LDC2008T19}), and the DUC 2001
corpus (\url{http://www-nlpir.nist.gov/projects/duc}). The first
dataset includes articles published in the NYTs over $3823$
days from 1996-2007.  
We collect the sentences in articles
associated with human generated summaries as the ground set $V$ (with sizes
varying from $2000$ to $20000$), and extract their TFIDF features to
build $f(S)$. We concatenate the sentences from all human generated
summaries for  the same date as a reference summary. We compare the
machine generated summaries produced by different methods with the
reference summary by ROUGE-2~\cite{rouge} (recall on 2-grams) and
ROUGE-2 F1-score (F1-measure based on recall and precision on
2-grams). 

We also compare their relative utility. As before,
sieve-streaming has memory set at $50k$. The statistics over
$3823$ days are shown in Figure~\ref{fig:NYT_utility}.
SS has a relative utility of $\geq 0.99$ on most days, while sieve-streaming is mostly in the $[0.92, 0.93]$ region. Both the ROUGE-2 and F1 score of SS are better than sieve-streaming, and even outperform greedy a bit. This may be because SS removes many of the elements on which greedy might become trapped in some local sub-optimal region.

\begin{wrapfigure}{r}{0.43\linewidth}
\vspace{-2em}
\begin{center}
 \includegraphics[width=1\linewidth]{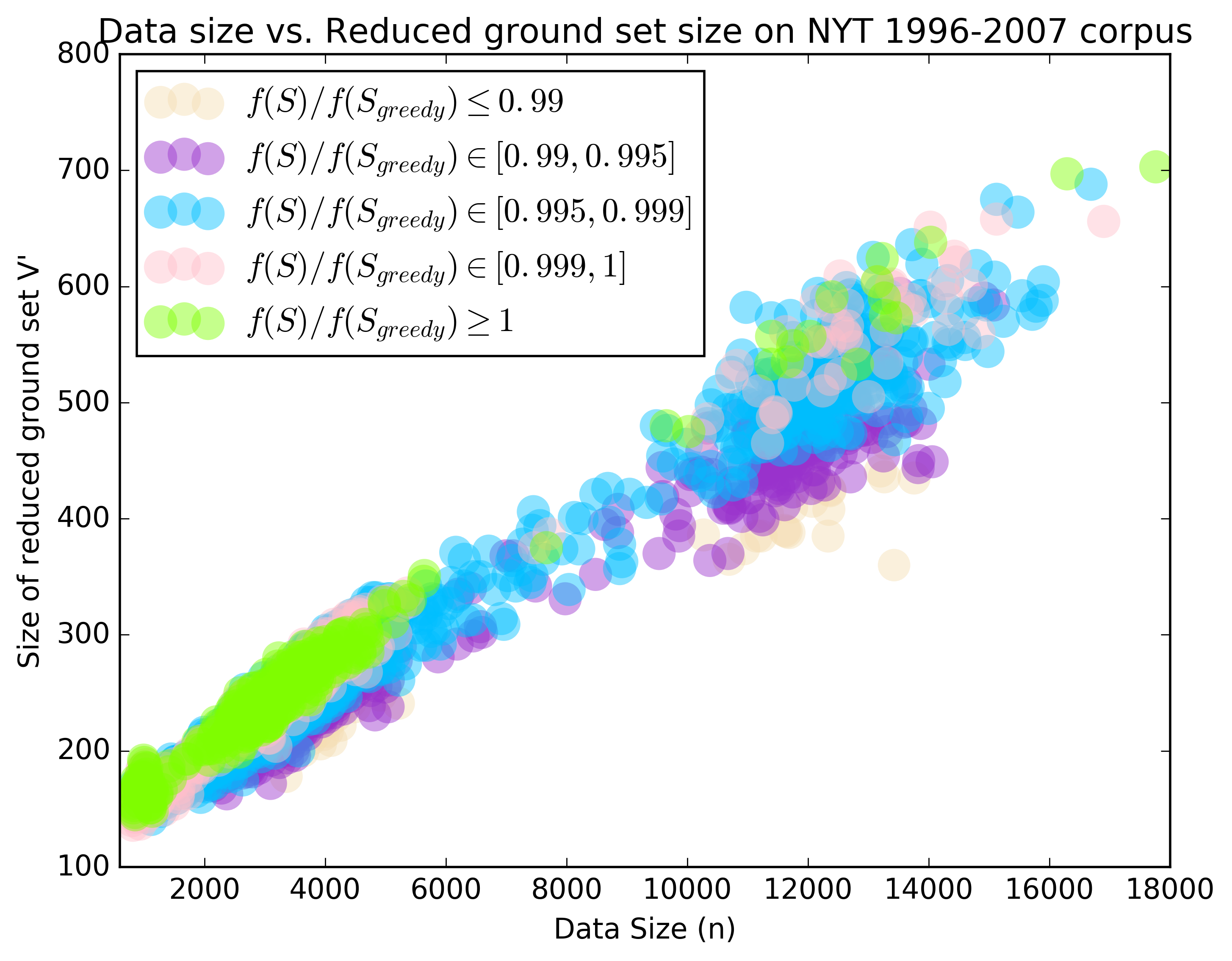}
\end{center}\vspace{-4mm}
   \caption{\scriptsize{Scatter plot of relative utility $f(S)/f(S_{greedy})$ achieved by submodular sparsification on the $3823$ days' news with the corresponding size of ground set $V$ and the size of reduced set $V'$. Each point corresponds to one day.}}
 \label{fig:NYT_mvn}
\vspace{-6mm}
\end{wrapfigure}
Figure~\ref{fig:NYT_nvstime} shows the number $n$ of sentences per day
and the corresponding time cost of each algorithm. The area of each
circle is proportional to relative utility. We use a log scale time
axis for a wider dynamic range. SS reduces computation over lazy
greedy especially when $n$ is large. Sieve-streaming's time cost
decreases when $n\geq 6000$, but its relative utility is reduced due
to the aforementioned early stopping.  Figure~\ref{fig:NYT_mvn} shows
the distribution of relative utility achieved by SS with different
data sizes $n$ and reduced ground set sizes over $3823$ different
days. The relative utility of SS is $\geq 0.99$ on most days, and even
$\geq 1$ when $n\leq 6000$. This indicates that summarization on the
reduced set $V'$ achieved by SS can even occasionally outperform that on the
original ground set $V$.

\vspace{-2mm}
\subsection{Video Summarization}
\label{sec:video-summarization}

We apply lazy greedy, sieve-streaming, and SS to $25$ videos from dataset SumMe \cite{Summe} (\url{http://www.vision.ee.ethz.ch/~gyglim/vsum/}). Each video has $1000\sim 10000$ frames as given in Table~\ref{table:SumMe} \cite{Supp}. The results are given in \cite{Supp}. The greedy algorithm on the SS-reduced ground set consistently approaches or outperforms lazy greedy on recall and F1-score, while the time cost is much smaller and a large fraction of frames may be removed.

%\vspace{-2mm}
%\section{Conclusion}
%
%We introduce ``submodularity graph (SG)'' describing pairwise relationship defined by a submodular objective for summarization. By using the graph properties, a random pruning method ``submodular sparsification (SS)'' is proposed to remove elements from the ground set for submodular maximization. We prove that SS can reduce the data size from $n$ to $O(\log^2 n)$ without causing too much loss on the objective. SS scales up submodular maximization to large $n$ on both memory and computation, while costs a few highly parallelizable computation. The reduced dataset approaches or even outperforms the original one in several summarization tasks on real-world data.

\newpage
\fontsize{9.5}{10}\selectfont
\setlength{\bibsep}{1pt}
\bibliography{sg}
\bibliographystyle{plain}
\normalsize

\newpage
\section{Appendix}
\label{sec:appendix}
\subsection{Proof of Lemma~\ref{lemma:triangle}}

\begin{proof}
	Firstly, we have the following inequality.
	\begin{align}\label{equ:triangle}
		\notag f(x|v)&=f(x+u|v)-f(u|v+x)\\
		\notag &=f(x|u+v)+f(u|v)-f(u|v+x)\\
		&\leq f(x|u)+f(u|v)-f(u|v+x).
	\end{align}
	The first two equalities follow from the definition of marginal gain, while the inequality is due to submodularity. Following the definition of $w_{uv}$ in Eq.~\eqref{equ:wuv}, we have
	\begin{align}
	\notag w_{vx}&=f(x|v)-f(v|V-v)\\
	\notag &\leq f(x|u)+f(u|v)-f(u|v+x)-f(v|V-v)\\
	\notag &\leq \left[f(x|u)-f(u|V- u)\right]+\left[f(u|v)-f(v|V-v)\right]\\
	&=w_{ux}+w_{vu}.
	\end{align}
	The first inequality is due to Eq.~\eqref{equ:triangle}, and the second inequality is via submodularity. 
\end{proof}

\subsection{Proof of Proposition~\ref{prop:sssm}}
\label{sec:proof-prop-refpr}

\begin{proof}
  Define a set $A_u$ for each $u\in V'$ such that $A_u=\{v\in V:
  w_{uv}\leq\epsilon\}$. Note $u\in A_u$ because
  $w_{uu}=-f(u|V\backslash u)\leq 0\leq\epsilon$ and
  hence $V' \subseteq \cup_{u \in V'} A_u$.
  The objective
  function $h$ in Eq.~\eqref{equ:problemss} can be written as
\begin{align}
h(V') 
&= \left|\left\{v\in V\backslash V': w_{V'v}\leq\epsilon\right\}\right|
= \left|\left\{v\in V\backslash V': \exists x \in V' : w_{xv}\leq\epsilon\right\}\right| \\
&=\left| \left( \bigcup_{u\in V'}A_u\right) \setminus V' \right|
=\left|\bigcup_{u\in V'}A_u\right|-\left|V'\right|,
\end{align}
where $f_{SC}(V')=\left|\bigcup_{u\in V'}A_u\right|$ is the simple set cover function \cite{Fujishige}, which is monotone non-decreasing submodular, and $-|V'|$ is a monotone decreasing modular (negative cardinality) function. Because the sum of a submodular function and a modular function is still submodular, the objective in Eq.~\eqref{equ:problemss} is non-monotone submodular.
\end{proof}

\subsection{Proof of Theorem~\ref{the:ssbound}}
\label{sec:proof-theor-refth}

\begin{proof}
Recall that $u^*_v \in \argmin_{u\in V^*}w_{uv}$ is the tail node of an edge with the minimal weight over all edges from elements in $V^*$ to head $v$. 
Since $|V^*|\geq k$, the greedy algorithm on $V^*$ will run for $k$ steps and select $k$ elements. We use $S_i$ to denote the solution set at the beginning of the $i^{th}$ step, let $u_i \in \argmax_{x\in V^* \setminus S_i}f(x|S_i)$ be the selected element in this step. In addition, let  $v_i=\argmax_{x\in V \setminus S_i}f(x|S_i)$ be the unfettered greedy choice at step $i$. Then 
we have the following:
\begin{equation}\label{equ:viui}
\begin{array}{ll}
f(v_i|S_i)&\leq f(u_i|S_i)+\min\limits_{x\in V^*}w_{xv_i|S}\\
&\leq f(u_i|S_i)+\min\limits_{x\in V^*}w_{xv_i}\\
&=f(u_i|S_i)+w_{u^*_{v_i}v_i}\\
&\leq f(u_i|S_i)+\epsilon.
\end{array}
\end{equation}
The first inequality is by Eq.~\eqref{equ:loss},
the second inequality is due to Lemma~\ref{lemma:wSwP}, while the last inequality comes from the definition of problem Eq.~\eqref{equ:problemss}. Hence, for arbitrary $i$, we have
\begin{equation}\label{equ:t21}
\begin{array}{ll}
f(S^*)&\leq f(S_i\cup S^*)\\
&\leq f(S_i)+\sum\limits_{x\in S^*\backslash S_i}f(x|S_i)\\
&\leq f(S_i)+\sum\limits_{x\in S^*}f(x|S_i)\\
&\leq f(S_i)+k\max\limits_{x\in V}f(x|S_i)\\
&=f(S_i)+kf(v_i|S_i)\\
&\leq f(S_i)+k\left[f(u_i|S_i)+\epsilon\right]\\
&= f(S_i)+k\left[f(S_{i+1})-f(S_i)+\epsilon\right].
\end{array}
\end{equation}
The first inequality uses monotonicity of $f(\cdot)$, while the second one is due to submodularity. The third inequality is due to the non-negativity of $f(\cdot)$. The fourth inequality is due to the maximal greedy selection rule for the greedy algorithm on the original ground set $V$. The fifth inequality is the result of applying Eq.~\eqref{equ:viui}. The last equality is due to the greedy selection rule $S_{i+1}=u_i\cup S_i$ for the greedy algorithm on the reduced ground set $V^*$. Rearranging Eq.~\eqref{equ:t21} yields
\begin{equation}
[f(S^*)-k\epsilon]-f(S_i)\leq k[f(S_{i+1})-f(S_i)]
\end{equation}
Let
\begin{equation}\label{equ:delta}
\delta_i=[f(S^*)-k\epsilon]-f(S_i),
\end{equation}
then the rearranged inequality equals to
\begin{equation}\label{equ:deltaj}
\delta_i\leq k[\delta_i-\delta_{i+1}],
\end{equation}
Since $\delta_i-\delta_{i+1} \geq 0$, this equals to
\begin{equation}\label{equ:jinequ}
\delta_{i+1}\leq\left(1-\frac{1}{k}\right)\delta_i.
\end{equation}
Since in total $k$ elements are selected by the greedy algorithm, applying Eq.~\eqref{equ:jinequ} from $i=0$ to $i=k$ yields
\begin{equation}
\delta_k\leq \left(1-\frac{1}{k}\right)^{k}\delta_0\leq e^{-1}\delta_0.
\end{equation}
By using the definition of $\delta_i$ in Eq.~\eqref{equ:delta}, the above inequality leads to
\begin{equation}
f(S')=f(S_k)\geq \left(1-e^{-1}\right)\left(f(S^*)-k\epsilon\right).
\end{equation}
This completes the proof.
\end{proof}

\subsection{Proof of Lemma~\ref{lemma:uv}}
\label{sec:proof-lemma-refl}

\begin{proof}
	The proof follows from Lemma~\ref{lemma:triangle} and our assumption to $u$.
	\begin{align}
	\notag w_{uv}\leq & w_{uu^*_v}+w_{u^*_vv}\\
	\notag = & f(v|u^*_v)+f(u^*_v|u)-f(u^*_v|V\backslash u^*_v)-f(u|V\backslash u)\\
	\notag = & f(v+u^*_v)+f(u+u^*_v)-f(u^*_v)-f(u)\\
	\notag &-f(u^*_v|V\backslash u^*_v)-f(u|V\backslash u)\\
	\notag \leq & 2f(v+u^*_v)-f(u^*_v)-f(u)\\
	\notag &-f(u^*_v|V\backslash u^*_v)-f(u|V\backslash u)\\
	\notag = & 2\left[f(v|u^*_v)-f(u^*_v|V\backslash u^*_v)\right]\\
	\notag & +\left[f(u^*_v)+f(u^*_v|V\backslash u^*_v)-f(u)-f(u|V\backslash u)\right]\\
	\notag \leq & 2w_{u^*_vv}.
	\end{align}
	The first inequality is due to Lemma~\ref{lemma:triangle}. The second inequality is because $f(u+u^*_v)\leq f(v+u^*_v)$ which follows from $u\in P(u^*_v)$. The third inequality is due to $u\in Q(u^*_v)$. 
\end{proof}

\subsection{Proof of Proposition~\ref{prop:outsideball}}
\label{sec:proof-prop-refpr-1}

\begin{proof}
	Recall $V^*$ is the optimal solution of problem in Eq.~\eqref{equ:problemss}. Due to the definition of $|V|/(8K)$-NN ball, we have
	\begin{equation}
	\forall v\in V_{u^*}\backslash B\left(u^*,|V|/(8K)\right), f(u+u^*)\leq f(v+u^*).
	\end{equation}
	Hence, $u\in P(u^*_v)\cap Q(u^*_v)$. By using Lemma~\ref{lemma:uv}, we have
	\begin{equation}
	w_{uv}\leq 2w_{u^*v}.
	\end{equation}
	This completes the proof.
\end{proof}

\subsection{Proof of Proposition~\ref{prop:upper4bad}}
\label{sec:proof-prop-refpr-3}

\begin{proof}
	According to Proposition~\ref{prop:outsideball}, for each $u^*\in V^*$, if one $u\in B\left(u^*,|V|/(cK)\right)\cap Q(u^*)$ is sampled into $U$ in some iteration of Algorithm~\ref{alg:coreset}, then any item $v$ outside the ball satisfies
		\begin{align}
		\notag w_{Uv}&=\min_{x\in U}w_{xv}\leq w_{uv}\\
		\notag &\leq 2w_{u^*_vv}=2w_{V^*v}.
		\end{align}
	Hence, one element $u$ fulfilling $w_{Uu}\geq 2w_{V^*u}$ in the complement set must be contained in least one of the $K$ $|V|/(cK)$-NN balls whose centers are the $K$ elements in $V^*$. Therefore, the total number of such $u$ is at most $|V|/c=K\times|V|/(cK)$, the maximal number of elements in all the $K$ $|V|/(cK)$-NN balls.
\end{proof}

\subsection{Proof of Proposition~\ref{prop:number_v}}
\label{sec:proof-prop-refpr-4}

\begin{proof}
	We consider $V_i$, set $V$ at the beginning of the $i^{th}$ iteration, and $V_{i-1}$, set $V$ right before the removal step of the previous iteration. According to the pruning amount $1-1/\sqrt{c}$:
	\begin{equation}
	|V_i|=1/\sqrt{c}|V_{i-1}|.
	\end{equation}
	Since Proposition~\ref{prop:upper4bad} indicates
	\begin{equation}
	\left|\{u\in V_i:w_{Uu}\geq 2w_{V^*u}\}\right|\leq \frac{|V_{i-1}|}{c},
	\end{equation}
	we have
	\begin{align}
	\notag &\left|\{v\in V_i:w_{Uv}\leq 2w_{V^*v}\}\right|\\
	\notag =&|V_i|-\left|\{u\in V_i:w_{Uu}\geq 2w_{V^*u}\}\right|\\
	\notag \geq& \frac{1}{\sqrt{c}}|V_{i-1}|-\frac{1}{c}|V_{i-1}|\\
	\notag =&\left(1-\frac{1}{\sqrt{c}}\right)\times(\frac{1}{\sqrt{c}})|V_{i-1}|\\
	\notag =&\left(1-\frac{1}{\sqrt{c}}\right)|V_i|.
	\end{align}

% \tianyi{Jeff}{TO ADD: for example, with $c=8$ we get a fast reduction rate of $\sqrt{2}/4$ with a pruning fraction of $1-\sqrt{2}/4$.}

	Because the above result is correct for arbitrary $i$, it completes the proof.
\end{proof}

\subsection{Proof of Lemma~\ref{lemma:condition}}
\label{sec:proof-lemma-refl-1}

\begin{proof}
According to Proposition~\ref{prop:number_v}, after removal, all the elements in $\{v\in V:w_{Uv}> 2w_{V^*v}\}$ are retained in $V'$. So none of them is in $V\backslash V'$. 

According to Proposition~\ref{prop:outsideball}, if for each $u^*\in V^*$ at least one alternate $u\in B\left(u^*,|V|/(cK)\right)\cap Q(u^*)$ is sampled and added into $U$, $\forall v\in V$, we have $w_{V'v}\leq 2w_{V^*v}$. This completes the proof.
\end{proof}

\subsection{Proof of Proposition~\ref{prop:upperpr}}
\label{sec:proof-prop-refpr-5}

\begin{proof}
	According to the assumption and definition of $Q(u^*)$ in Lemma~\ref{lemma:uv}, $\forall u\in U$,
	\begin{equation}\label{equ:rho}
	\Pr\left(u\in Q(u^*)\right)\geq q.
	\end{equation}
	In addition, the probability for that an uniform sample $u$ is inside the $|V|/(cK)$-NN ball $B\left(u^*,|V|/(cK)\right)$ of $u^*$ is
	\begin{equation}
	\Pr\left(u\in B\left(u^*,|V|/(cK)\right)\right)=\frac{1}{cK}.
	\end{equation}
	Combining the two probabilities, we have
	\begin{equation}
	\Pr\left(u\not\in B\left(u^*,|V|/(cK)\right)\cap Q(u^*)\right)\leq 1-\frac{q}{cK}.
	\end{equation}
	Since $r=O(cK)=pcK$, among the $r\log n=pcK\log n$ samples of $U$ in one iteration, for one specific $u^*$, the probability that no sample belongs to $B\left(u^*,|V|/(8K)\right)\cap Q(u^*)$ is
	\begin{align}
	\notag \Pr&\left(U\cap \left(B\left(u^*,|V|/(8K)\right)\cap Q(u^*)\right)=\emptyset\right)\\
	\notag &\leq\left(1-\frac{q}{cK}\right)^{r}=\left(1-\frac{q}{cK}\right)^{pcK\log n}\leq n^{-q p}.
	\end{align}
	Note there are $K$ items in $V^*$, and there will be at most $\log_{\sqrt{c}} n$ iterations. By union bound, the failure probability that no $u\in B(u^*,|V|/(cK))\cap Q(u^*)$ is sampled and added into $U$ for at least one $u^*\in V^*$ in at least one iteration of Algorithm~\ref{alg:coreset} is at most
	\begin{equation}
	K\times n^{-q p}\times \log_{\sqrt{c}} n\leq n^{1-q p}\log_{\sqrt{c}} n.
	\end{equation}
\end{proof}

\subsection{Proof of Theorem~\ref{the:ssbound}}
\label{sec:proof-theor-refth-1}

\begin{proof}
Firstly, since $r\log n=pcK\log n$ elements are selected into $V'$ per iteration, and the number of iterations is $\log_{\sqrt{c}} n$, so the size of $V'$ is
\begin{equation}
|V'|=pcK\log n\times \log_{\sqrt{c}} n=(pc/\log_{\sqrt{c}})K\log^2 n.
\end{equation}
Secondly, combing the results of Lemma~\ref{lemma:condition} and failure probability $n^{1-q p}\log_{\sqrt{c}} n$ in Proposition~\ref{prop:upperpr}, we have: with success probability $1-n^{1-q p}\log_{\sqrt{c}} n$, $\forall v\in V\backslash V'$, $w_{V'v}\leq 2w_{V^*v}$.

Thirdly, since $w_{V'v}\leq 2w_{V^*v}$, we replace $w_{u^*_{v_i}v_i}$ with $2w_{u^*_{v_i}v_i}$ in Eq.~\eqref{equ:viui}, the rest proof of Theorem~\ref{the:ssbound} leads to
\begin{equation}
f(S')\geq \left(1-e^{-1}\right)\left(f(S^*)-2k\epsilon\right).
\end{equation}
This completes the proof.
\end{proof}

\subsection{Experiments on DUC2001 News Summarization}
\label{sec:exper-duc2001-news}

\begin{figure}[htp]\vspace{-1mm}
 \begin{center}
  \includegraphics[width=0.65\linewidth]{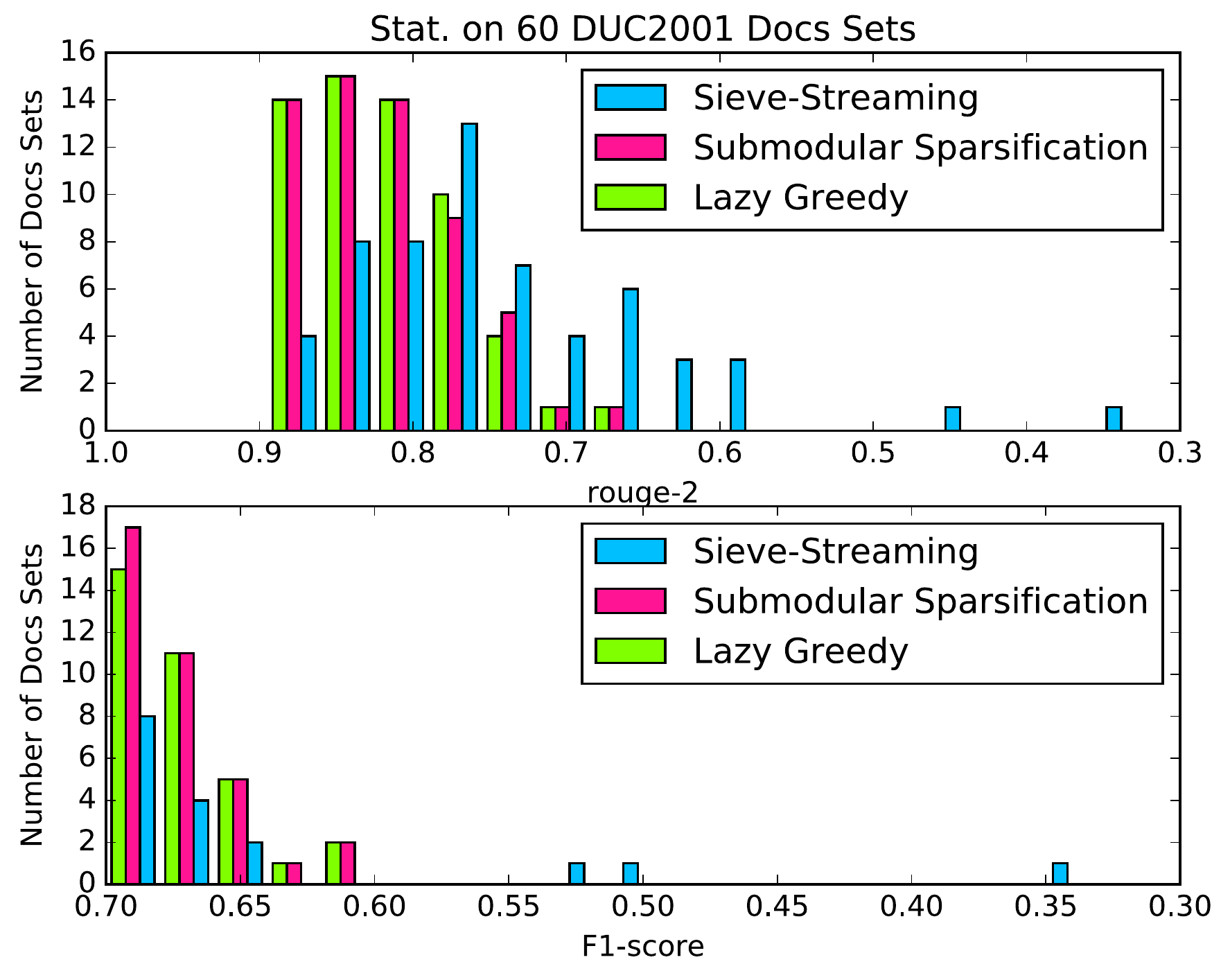}
%  \includegraphics[width=1\linewidth]{DUC2001_400_utility.png}
%\vspace{.5in}
 \end{center}\vspace{-4mm}
   \caption{Statistics of relative utility $f(S)/f(S_{greedy})$, ROUGE-2 score and F1-score on topic based news summarization results of $60$ document sets from DUC2001 training and test set, comparing to $400$-word human generated summary.}
 \label{fig:DUC2001_400_utility}
\end{figure}

\begin{figure}[htp]\vspace{-1mm}
 \begin{center}
  \includegraphics[width=0.65\linewidth]{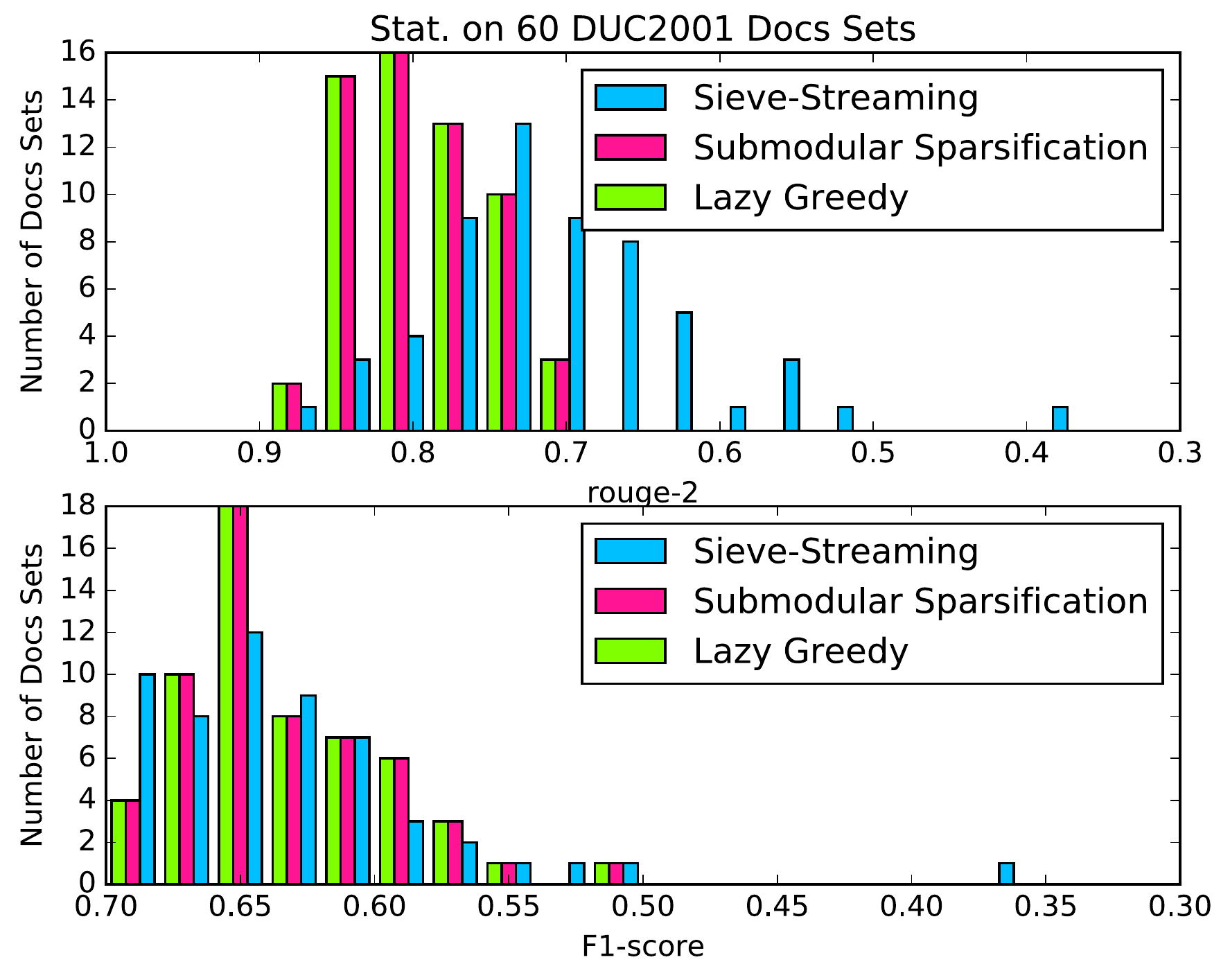}
%  \includegraphics[width=1\linewidth]{DUC2001_200_utility.png}
%\vspace{.5in}
 \end{center}\vspace{-4mm}
   \caption{Statistics of relative utility $f(S)/f(S_{greedy})$, ROUGE-2 score and F1-score on topic based news summarization results of $60$ document sets from DUC2001 training and test set, comparing to $200$-word human generated summary.}
 \label{fig:DUC2001_200_utility}
\end{figure}

We also observe similar result on DUC 2001 corpus, which are composed of two datasets. The first one includes $60$ sets of documents, each is selected by a NIST assessor because the documents in a set are related to a same topic. The assessor also provides four human generated summary of word count $400,200,100,50$ for each set. In Figure~\ref{fig:DUC2001_400_utility} and Figure~\ref{fig:DUC2001_200_utility}, we report the statistics to ROUGE-2 and F1-score of summaries of the same size generated by different algorithms. The second dataset is composed of four document sets associated with four topics. We report the detailed results in Table~\ref{table:DUC01}. Both of them show submodular sparsification can achieve similar performance as greedy algorithm, whereas outperforms sieve-streaming.

\vspace{-1mm}

\vspace{-1mm}

 \vspace{-1mm}
 \begin{table*}[htp]
 \caption{Performance of Lazy greed, sieve-streaming, and submodular sparsification on four topic summarization datasets from DUC 2001. For each topic, the machine generated summary is compared to four human generated ones of word count from 50 to 400.}
 \begin{center}\scriptsize
 \begin{tabular}{l|l|cc|cc|cc|cc}
 \hline
 \multirow{2}{*}{Algorithm} &\multirow{2}{*}{words} &\multicolumn{2}{c|}{Daycare} &\multicolumn{2}{c|}{Healthcare} &\multicolumn{2}{c|}{Pres92} & \multicolumn{2}{c}{Robert Gates}\\
 \cline{3-10}
 & &ROUGE2 &F1 &ROUGE2 &F1 &ROUGE2 &F1 &ROUGE2 &F1\\
 \hline
 \multirow{4}{*}{Lazy Greedy} &400 &$0.836$ &$0.674$ &$0.845$ &$0.686$ &$0.885$ &$0.686$ &$0.849$ &$0.734$\\
 &200 &$0.813$ &$0.615$ &$0.811$ &$0.632$ &$0.842$ &$0.623$ &$0.788$ &$0.682$\\
 &100 &$0.766$ &$0.542$ &$0.753$ &$0.605$ &$0.618$ &$0.420$ &$0.715$ &$0.621$\\
 &50 &$0.674$ &$0.484$ &$0.765$ &$0.539$ &$0.602$ &$0.341$ &$0.631$ &$0.514$\\
 \hline
 \multirow{4}{*}{Sieve-Streaming} &400 &$0.825$ &$0.687$ &$0.814$ &$0.711$ &$0.827$ &$0.710$ &$0.798$ &$0.745$\\
 &200 &$0.789$ &$0.627$ &$0.782$ &$0.675$ &$0.670$ &$0.659$ &$0.691$ &$0.688$\\
 &100 &$0.747$ &$0.542$ &$0.658$ &$0.597$ &$0.414$ &$0.443$ &$0.632$ &$0.620$\\
 &50 &$0.607$ &$0.475$ &$0.681$ &$0.551$ &$0.413$ &$0.345$ &$0.553$ &$0.477$\\
 \hline
 \multirow{4}{*}{SS}  &400 &$0.837$ &$0.674$ &$0.845$ &$0.686$ &$0.883$ &$0.685$ &$0.849$ &$0.734$\\
 &200 &$0.813$ &$0.615$ &$0.811$ &$0.632$ &$0.842$ &$0.623$ &$0.788$ &$0.682$\\
 &100 &$0.766$ &$0.542$ &$0.753$ &$0.605$ &$0.617$ &$0.420$ &$0.715$ &$0.621$\\
 &50 &$0.674$ &$0.484$ &$0.765$ &$0.539$ &$0.602$ &$0.341$ &$0.631$ &$0.514$\\
 \hline
 \end{tabular}\label{table:DUC01}
 \end{center}\vspace{-2mm}
 \vspace{-2mm}
 \end{table*}
 \normalsize

\subsection{Experiments on Video Summarization}
\label{sec:exper-video-summ}

\subsection{Video Summarization}
\label{sec:video-summarization-1}

We apply lazy greedy, sieve-streaming, and SS to $25$ videos from video summarization dataset SumMe \cite{Summe}\footnote{http://www.vision.ee.ethz.ch/$\sim$gyglim/vsum/}. Each video has $1000\sim 10000$ frames as given in Table~\ref{table:SumMe}.

We resize each frame to a $180\times 360$ image, and extract features from two standard image descriptors, i.e., a pyramid of HoG (pHoG) \cite{pHoG} to delineate local and global shape, and GIST \cite{GIST} to capture global scene. The $2728$ pHoG features are achieved over a four-level pyramid using $8$ bins with angle of $360$ degrees. The $256$ GIST features are obtained by using $4\times 4$ blocks and $8$ orientation per scale. We concatenate them to form a $2984$-dimensional feature vector for each frame to build $f(\cdot)$. Each algorithm selects $15\%$ of all frames as summary set, i.e., $k=0.15|V|$. Sieve-streaming holds a memory of $10k$ frames. 

\begin{figure*}[htp]\vspace{-3mm}
  \begin{center}
   \includegraphics[width=1\linewidth]{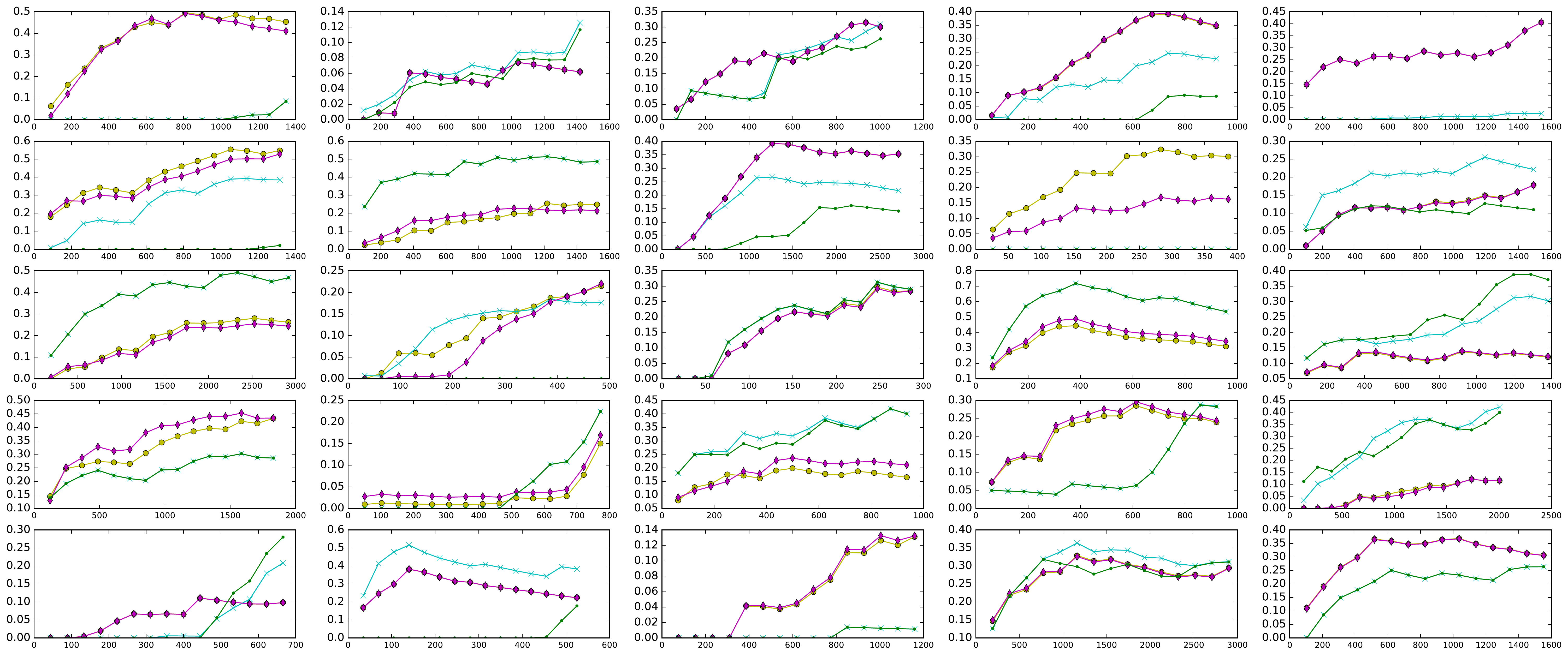}
 %  \includegraphics[width=1\linewidth]{DUC2001_200_utility.png}
 %\vspace{.5in}
  \end{center}\vspace{-5mm}
    \caption{F1-score of the summaries generated by lazy greedy (``{\color{yellow}{$\bullet$}}''), sieve-streaming ( ``{\color{cyan}{$\times$}}''), submodular sparsification (``{\color{magenta}{$\blacklozenge$}}'') and the first $15\%$ frames (``{\color{green}{$\cdot$}}'') comparing to reference summaries of different sizes between $[0.02|V|, 0.32|V|]$ based on ground truth score (voting from $15$ users) on $25$ videos from SumMe. Each plot associates with a video.}
  \label{fig:videoF1}
\end{figure*}

\begin{figure*}[htp]\vspace{-3mm}
  \begin{center}
   \includegraphics[width=1\linewidth]{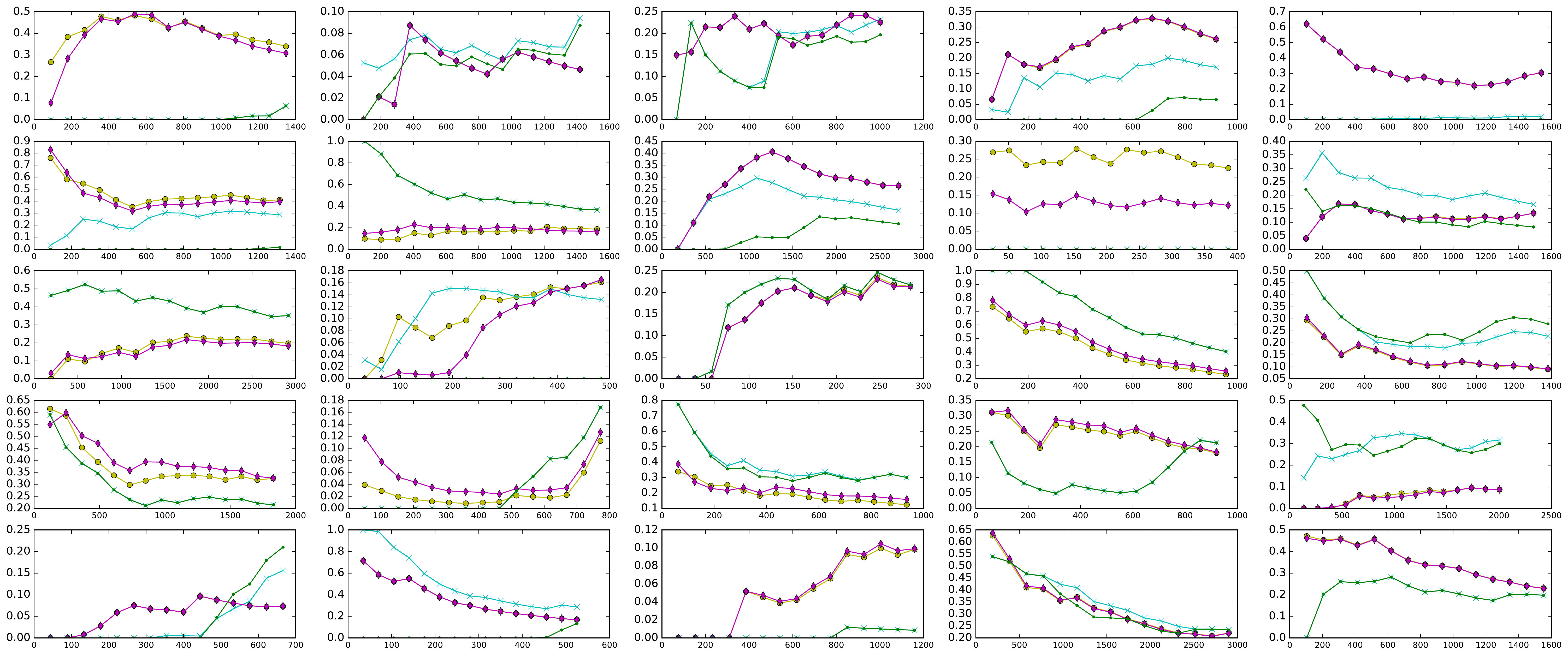}
 %  \includegraphics[width=1\linewidth]{DUC2001_200_utility.png}
 %\vspace{.5in}
  \end{center}\vspace{-5mm}
    \caption{Recall of the summaries generated by lazy greedy (``{\color{yellow}{$\bullet$}}''), sieve-streaming ( ``{\color{cyan}{$\times$}}''), submodular sparsification (``{\color{magenta}{$\blacklozenge$}}'') and the first $15\%$ frames (``{\color{green}{$\cdot$}}'') comparing to reference summaries of different sizes between $[0.02|V|, 0.32|V|]$ based on ground truth score (voting from $15$ users) on $25$ videos from SumMe. Each plot associates with a video.}
  \label{fig:videoR}
\end{figure*}

\begin{figure*}[htp]\vspace{-1mm}
  \begin{center}
   \includegraphics[width=1\linewidth]{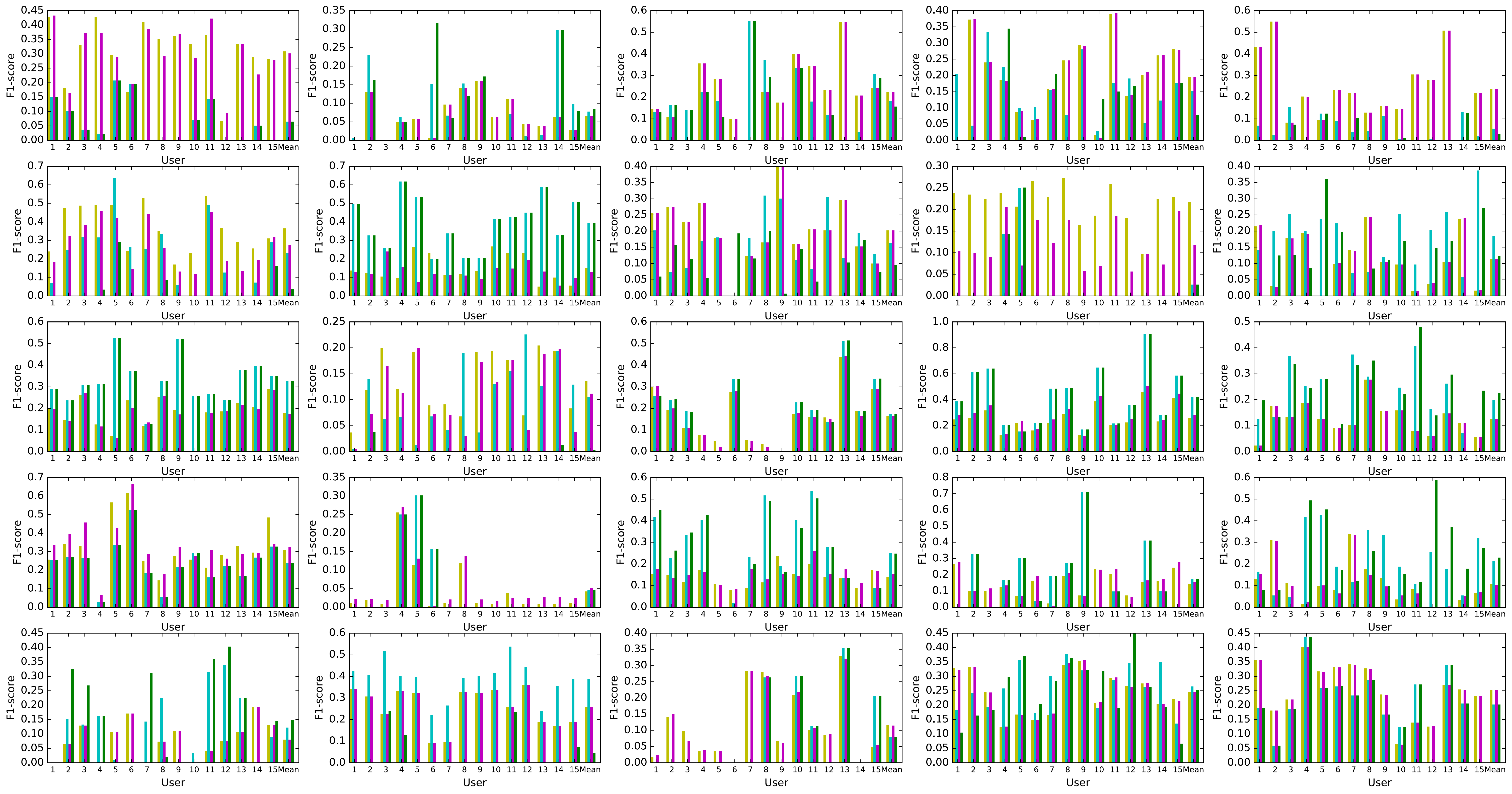}
 %  \includegraphics[width=1\linewidth]{DUC2001_200_utility.png}
 %\vspace{.5in}
  \end{center}\vspace{-2mm}
    \caption{F1-score of the summaries generated by greedy (yellow bar), sieve-streaming ( cyan bar), SS (magenta bar) and the first $15\%$ frames (green bar) comparing to reference summaries from $15$ users on $25$ videos from SumMe dataset. Each plot associates with a video.}
  \label{fig:videoF1_all}
\end{figure*}

\begin{figure*}[htp]\vspace{-1mm}
  \begin{center}
   \includegraphics[width=1\linewidth]{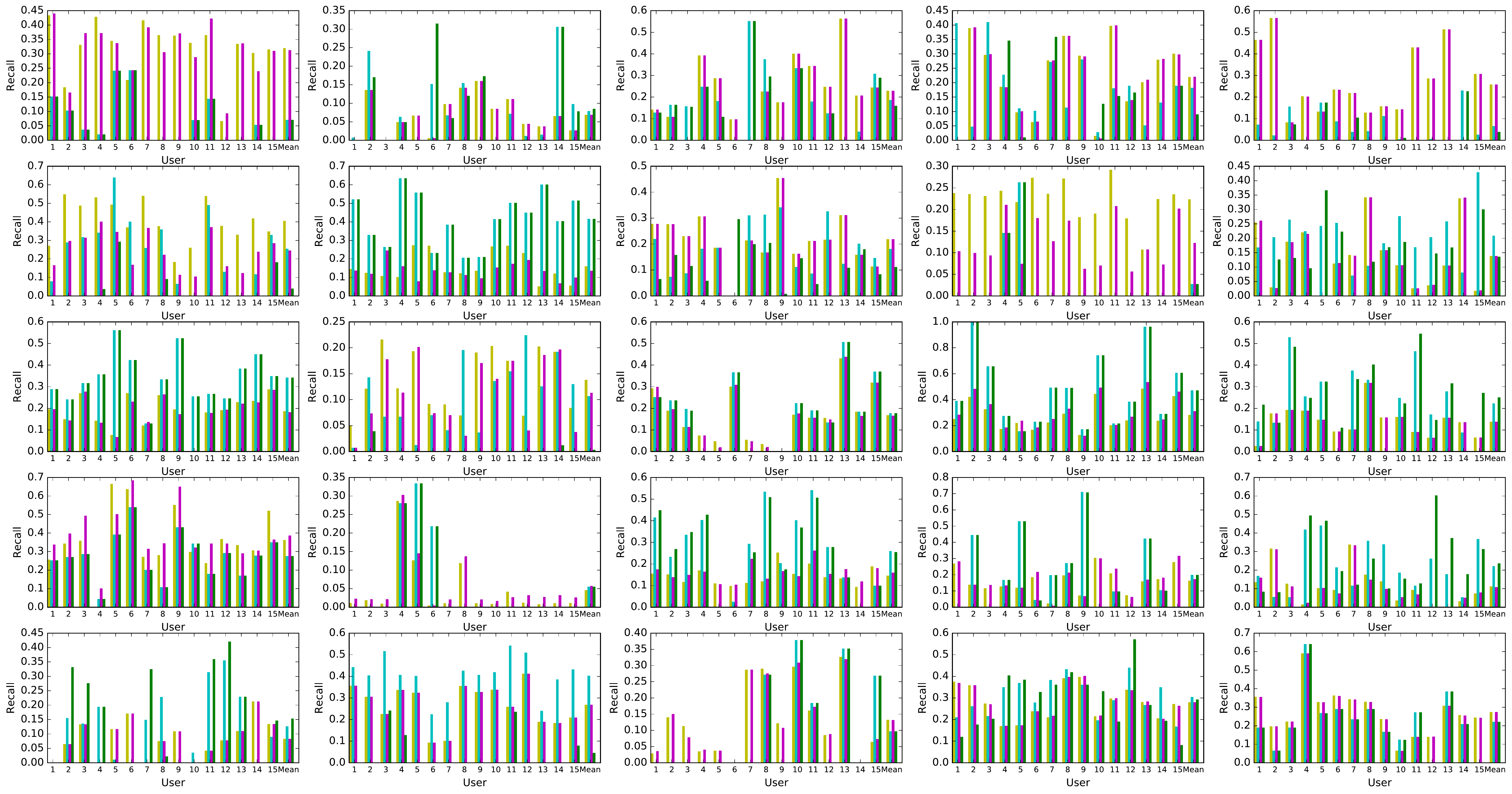}
 %  \includegraphics[width=1\linewidth]{DUC2001_200_utility.png}
 %\vspace{.5in}
  \end{center}\vspace{-2mm}
    \caption{Recall of the summaries generated by greedy (yellow bar), sieve-streaming ( cyan bar), SS (magenta bar) and the first $15\%$ frames (green bar) comparing to reference summaries from $15$ users on $25$ videos from SumMe dataset. Each plot associates with a video.}
  \label{fig:videoR_all}
\end{figure*}

We compare the summaries generated by the three algorithms with the ones produced by the ground truth and $15$ users. Each user was asked to select a subset of frames as summary, and ground truth score of each frame is given by voting from all $15$ users. For each video, we compare each algorithm generated summary with the reference summary composed of the top $p$ frames with the largest ground truth scores for different $p$, and the user summary from different users. In particular, we report F1-score and recall for comparison to ground truth score generated summaries in Figure~\ref{fig:videoF1} and Figure~\ref{fig:videoR}. We report F1-score and recall for comparison to user summaries in Figure~\ref{fig:videoF1_all} and Figure~\ref{fig:videoR_all}. In each plot for each video, we also report the average F1-score and average recall over all $15$ users.

SS consistently approaches or outperforms lazy greedy, while the time cost is much smaller according to Table~\ref{table:SumMe} \cite{Supp}. Although on a few videos sieve-streaming achieves the best F1-score, in these cases its generated summaries are trivially dominated by the first $15\%$ frames as shown in Figure~\ref{fig:videoF1}-\ref{fig:videoR_all}.

\vspace{-3mm}

\vspace{-3mm}

\vspace{-1mm}

\vspace{-1mm}

\vspace{-1mm}
 \begin{table*}[htp]
 \caption{Information of SumMe dataset and time cost (CPU seconds) of different algorithms.}
 \begin{center}\small
 \begin{tabular}{l|c|c||c|c|c}
 \hline
 Video &\#frames & $|V'|$ &Lazy Greedy &Sieve-streaming &SS\\
 \hline
 Air Force One &4494 &1031 &907.3712 &3.9182  &71.4521 \\
 Base jumping &4729 &1074 &164.1434 &5.5865  &84.6877 \\
 Bearpark climbing &3341 &1038 &177.8583 &3.7311  &48.0415 \\
 Bike polo &3064 &866 &96.5305 &3.9578  &36.4832 \\
 Bus in rock tunnel &5131 &1387 &505.7766 &6.0088  &125.8121 \\
 Car over camera &4382 &1396 &146.9416 &5.3323  &69.6157 \\
 Car railcrossing &5075 &1210 &852.1686 &5.2265 &96.2396 \\
 Cockpit landing &9046 &2292 &669.8063 &12.3186 &212.7866 \\
 Cooking &1286 &200 &30.0717 &1.2868 &5.7096 \\
 Eiffel tower &4971 &1647 &304.2690 &5.4755 &86.5552\\
 Excavators river crossing &9721 &1971 &1507.3028 &13.8139 &284.5136 \\
 Fire Domino &1612 &464 &34.2871 &1.8814 &9.9833 \\
 Jumps &950 &308 &15.0508 &0.9055 &4.8719 \\
 Kids playing in leaves &3187 &986 &221.4644 &3.4660 &41.1956 \\
 Notre Dame &4608 &1136 &169.1235 &5.1406 &72.9076 \\
 Paintball &6096 &1664 &763.3255 &6.7853 &128.1723 \\
 Paluma jump &2574 &727 &210.8670 &2.5342 &26.7430 \\
 Playing ball &3120 &697 &132.7437 &3.2250 &32.3198\\
 Playing on water slide &3065 &778 &111.7358 &3.4088 &30.4131 \\
 Saving dolphines &6683 &1860 &435.0732 &7.3322 &121.5891\\
 Scuba &2221 &775 &45.6177 &2.5213 &18.4227 \\
 St Maarten Landing &1751 &628 &19.0717 &2.8701 &12.4074 \\
 Statue of Liberty &3863 &1223 &160.7075 &4.0164 &55.7420 \\
 Uncut evening flight &9672 &3324 &718.7015 &14.6717 &208.8540 \\
 Valparaiso downhill &5178 &1438 &428.3941 &6.0002 &154.5902\\
 \hline
 \end{tabular}\label{table:SumMe}
 \end{center}\vspace{-2mm}
 \vspace{-2mm}
 \end{table*}
 \normalsize

\end{document}